\documentclass[twoside,11pt]{article}
\usepackage{epsfig,graphicx,color,dsfont}
\usepackage[english]{babel}
\usepackage[latin1]{inputenc}
\usepackage{amsmath,amsfonts, amssymb   }

\newcommand{\BR}{BR}
\newcommand{\diag}{diag}

\newcommand{\NC}{NC}

\begin{document}

\newcounter{compteur}
\newtheorem{proposition}{Proposition}[section]
\newtheorem{theorem}[proposition]{Theorem}
\newtheorem{lemma}[proposition]{Lemma}
\newtheorem{corollary}[proposition]{Corollary}
\newtheorem{definition}[proposition]{Definition}
\newtheorem{remark}[proposition]{Remark}
\newtheorem{example}[proposition]{Example}

\title{Internal Regret with Partial Monitoring\\ Calibration-Based Optimal Algorithms}

\author{ Vianney Perchet\thanks{Centre de Math\'ematiques et de Leurs Applications UMR 8536, \'Ecole Normale Supérieure, 61, avenue du pr\'esident Wilson, 94235 Cachan, France. vianney.perchet@normalesup.}}

\maketitle

\begin{abstract}
We provide consistent random algorithms for sequential decision under partial monitoring, \textsl{i.e.}\ when the
decision maker does not observe the outcomes but receives instead
random feedback signals. Those algorithms have no internal regret in the sense that,
 on the set of stages  where the decision maker chose
his action according to a given law, the average payoff could not
have been improved in average by using any other fixed law.

They are based on a generalization of calibration, no longer defined in terms of  a Vorono\"i diagram but instead of
 a Laguerre diagram (a more general concept). This allows us to bound, for the first time in this general framework,  the expected
average internal -- as well as the usual external -- regret at stage $n$ by $O(n^{-1/3})$, which is known to be optimal.

\bigskip

\textsl{Key Words :}
Repeated games, On-line learning, Regret, Partial Monitoring, Calibration, Vorono\"i and Laguerre Diagrams
\end{abstract}

Hannan \cite{Hannan} introduced the notion of regret   in repeated games: a
player (that will be referred as a decision maker or also  a
forecaster) has no external regret if, asymptotically, his average
payoff could not have been greater if he had known, before the
beginning of the game, the empirical distribution of moves of the
other player. Blackwell \cite{BlackwellControlled} showed that the existence of such \textsl{externally consistent}
strategies, first proved by \cite{Hannan}, is a consequence of his approachability
theorem. A generalization of this result and   a more precise notion
of regret are due to Foster \& Vohra
\cite{FosterVohraCalibratedLearningCorrelatedEquilibrium} and Fudenberg \& Levine
\cite{FudenbergLevineConditionalUniversalConsistency}: there exist
internally consistent strategies, \textsl{i.e.}\ such that for any of
his action, the  decision maker has no external regret on the set of
stages where he actually chose this specific action. Hart \& Mas-Colell
\cite{HartMasColellCorrelatedEquilibrium} also used Blackwell's
approachability theorem to construct explicit algorithms that bound
 the internal (and therefore the external) regret at stage
$n$  by $O\left(n^{-1/2}\right)$.

\medskip

Some of those results have been extended to the partial monitoring
framework, \textsl{i.e.}\ where the decision maker  receives  at each
stage a random signal, whose law might depend on his unobserved payoff. Rustichini \cite{Rustichini} defined - and proved the existence
of - externally consistent strategies, \textsl{i.e.}\ such that the
average payoff of the decision maker could not have been
asymptotically greater if he had known, before the beginning of the
game, the empirical distribution of signals. Actually, the relevant
information is a vector of probability distributions, one for each
action of the decision maker,
that is called \textsl{a flag}.

Some algorithms bounding optimally the expected regret by
$O\left(n^{-1/3}\right)$ have been exhibited under some strong assumptions on the signalling structure -- see
Cesa-Bianchi \& Lugosi \cite{CesaBianchiLugosi}, Theorem 6.7 for the optimality of this bound. For example, Jaksch, Ortner \& Auer \cite{JakschOrtnerAuer} considered the Markov decision process framework, Cesa-Bianchi, Lugosi \& Stoltz \cite{CesaBianchiLugosiStoltz} assumed that payoffs can be deduced from flags and  Lugosi, Mannor \& Stoltz \cite{LugosiMannorStoltz} that feedbacks are deterministic (along with the fact that the worst compatible payoff is linear with respect to the flag). When
no such assumption is made, Lugosi, Mannor \& Stoltz \cite{LugosiMannorStoltz} provided an algorithm (based on the
exponential  weight algorithm) that bounds  regret by $O\left(n^{-1/5}\right)$.

\bigskip

In this framework, internal regret  was defined by Lehrer \& Solan
\cite{LehrerSolanPSE};  stages are no longer distinguished as a
function of the action chosen by the decision maker  (as in the full
monitoring case) but as a function of its law. Indeed, the
evaluation of the  payoff  (usually called \textsl{worst case}) is
not linear with respect to the flag. So  a best response - in a
sense to be defined - to a given flag might consist only in  a mixed
action (\textsl{i.e.}\ a probability distribution over the set of
actions).  Lehrer \& Solan \cite{LehrerSolanPSE} also proved the existence and constructed
internally consistent strategies, using the characterization of approachable convex sets due to Blackwell \cite{BlackwellAnalogue}. Perchet \cite{PerchetCalibrationALT}
provided an alternative  algorithm, recalled in section~\ref{sectionnaivealgo}; this latter is  based on calibration, a notion introduced
by Dawid \cite{DawidWellCalibrated}. Roughly speaking, these algorithms
$\varepsilon$-discretize arbitrarily the space of flags and each point of the discretization is called a possible
prediction. Then, stage after stage, they predict what will be the
next flag and output a \textsl{best response} to it. If the
sequence of predictions is calibrated then the average flag, on the
set of stages where  a specific prediction is made, will be close to
this prediction.

Thanks to the continuity of payoff and signaling functions, both algorithms bound  the internal regret by $\varepsilon + O\left(n^{-1/2}\right)$. However the first drawback lies in their computational complexities: at each stage, the algorithm of Perchet \cite{PerchetCalibrationALT} solves a system of linear equations  while the one  Lehrer \& Solan \cite{LehrerSolanPSE}, after a projection on a convex set, solves  a linear program. In both case, the size of the linear system or  program considered is polynomial in $\varepsilon$ and exponential in the numbers of actions and signals. The second drawback  is that the constants in the rate of convergence depend drastically on $\varepsilon$.

As a consequence, a classic \textsl{doubling trick} argument will generate an algorithm with a strongly sub-optimal rate of convergence -- that  might even depend on the size of the actions sets -- and a complexity that increases  with time.

\medskip

Our main  result  is Theorem \ref{theoaction}, stated in section \ref{optimal}: it provides the first algorithm that bounds optimally both internal and external regret by $O\left(n^{-1/3}\right)$ in the general case. It is a modification of the algorithm of Perchet \cite{PerchetCalibrationALT} that does not use an arbitrary
discretization but constructs carefully a specific one and then
computes, stage by stage, the solution of a system of linear
equations of constant size. In section \ref{optimal2}, an other
algorithm -- based on Blackwell's approachability as the one of  Lehrer \& Solan \cite{LehrerSolanPSE} -- with optimal rate and smaller constants is exhibited; it requires however to solve, at each stage,  a linear program of constant size.

\bigskip

Section 1 is devoted to the simpler framework of full monitoring. We recall
definitions of calibration and regret and we provide a na\"ive
algorithm  to construct strategies with internal regret asymptotically smaller
than $\varepsilon$. We show how to modify this algorithm -- however in a not efficient way -- in order to
bound optimally the regret by $O\left(n^{-1/2}\right)$. This has to be seen only as a tool that can be easily adapted with partial monitoring in order to reach the optimal bound
of $O\left(n^{-1/3}\right)$;  this is done in section 2. Some extensions (the second algorithm,
the so-called \textsl{compact case} and  variants to strengthen
the constants) are presented in section 3. Some technical proofs can be found in Appendix.

\section{Full monitoring}
\subsection{Model and definitions}
Consider a two-person game $\Gamma$ repeated  in discrete time, where at
stage $n \in \mathbb{N}$, a decision maker, or forecaster, (resp.\
the environment or Nature) chooses an action $i_n \in \mathcal{I}$ (resp.\
$j_n \in \mathcal{J}$). This generates a
payoff $\rho_n=\rho(i_n,j_n)$, where $\rho$ is a mapping from  $\mathcal{I}
\times \mathcal{J}$ to $\mathbb{R}$, and  a regret $r_n \in \mathbb{R}^{I}$
defined by:
\[r_n =\Big[\rho(i,j_n)-\rho(i_n,j_n)\Big]_{i \in \mathcal{I}} \in \mathbb{R}^{I},\]
where $I$ is the finite cardinality of $\mathcal{I}$ (and  $J$ the one of $\mathcal{J}$). This vector represents the differences between what the decision
maker could have got and what he actually got.

The choices  of $i_n$ and $j_n$ depend on
  the past observations (also called finite history)
$h_{n-1}=\left(i_1,j_1,\ldots,i_{n-1},j_{n-1}\right)$ and may be
random. Explicitly, the set of finite histories is denoted by $H =
\bigcup_{n \in \mathbb{N}} \left(\mathcal{I} \times \mathcal{J}\right)^n$, with $\left(\mathcal{I}
\times \mathcal{J}\right)^0=\emptyset$ and a strategy $\sigma$ of the decision
maker  is a mapping from $H$ to $\Delta(\mathcal{I})$, the set of probability distributions over $\mathcal{I}$. Given
the history $h_n \in \left(\mathcal{I} \times \mathcal{J}\right)^n$, $\sigma(h_n) \in \Delta(\mathcal{I})$ is
the law of $i_{n+1}$. A  strategy $\tau$ of Nature is defined
similarly as a function from $H$ to $\Delta(\mathcal{J})$.  A pair of strategies $(\sigma,\tau)$ generates a
probability, denoted by $\mathbb{P}_{\sigma,\tau}$, over
$(\mathcal{H}, \mathcal{A})$ where $\mathcal{H}=\left(\mathcal{I }\times \mathcal{J}\right)^{\mathbb{N}}$ is the set of
infinite histories embedded with the cylinder $\sigma$-field.

\bigskip

We extend the payoff mapping $\rho$  to $\Delta(\mathcal{I}) \times
\Delta(\mathcal{J})$ by $\rho(x,y)=\mathbb{E}_{x,y}[\rho(i,j)]$ and  for any sequence $a=\left(a_m\right)_{m \in
\mathbb{N}}$ and any $n \in \mathbb{N}_*$, we denote by
$\bar{a}_n=\frac{1}{n}\sum_{m=1}^na_m$  the average of $a$ up to stage
$n$.

\begin{definition}[Hannan \cite{Hannan}]
A strategy $\sigma$ of the forecaster is
externally consistent if for every strategy $\tau$ of
Nature:
\[\limsup_{n \to \infty} \bar{r}^i_n \leq 0, \quad \forall i \in \mathcal{I}, \quad \mathbb{P}_{\sigma,\tau}\mathrm{-as}.\]
\end{definition}
In words, a strategy $\sigma$ is externally consistent
if the forecaster  could not have had a greater payoff if he had
known, before the beginning of the game, the empirical distribution
of actions of Nature. Indeed, the external consistency of $\sigma$ is equivalent to the fact that :
\begin{equation}\label{equadefeqregrext}\limsup_{n \to \infty} \max_{x \in \Delta(\mathcal{I})} \rho(x,\bar{\jmath}_n)-\bar{\rho}_n \leq 0,  \quad \mathbb{P}_{\sigma,\tau}\mathrm{-as}.\end{equation}

Foster \& Vohra \cite{FosterVohraCalibratedLearningCorrelatedEquilibrium} (see also Fudenberg \& Levine
\cite{FudenbergLevineConditionalUniversalConsistency}) defined a more precise notion of regret.
 The internal regret of the stage $n$, denoted by $R_n \in \mathbb{R}^{I\times I}$, is also generated by the  choices of $i_n$ and $j_n$ and its $(i,k)$-th coordinate  is defined by:
\[R_n^{ik}=\left\{\begin{array}{cc} \rho(k,j_n)-\rho(i,j_n) & \mathrm{if\ } i=i_n\\
0& \mathrm{otherwise.}\end{array}\right.\]
Stated differently, every row of the matrix $R_n$  is null except the $i_n$-th which is $r_n$.
\begin{definition}[Foster \& Vohra \cite{FosterVohraCalibratedLearningCorrelatedEquilibrium}]\label{defintregret}
A strategy $\sigma$ of the forecaster is
internally consistent if  for every strategy $\tau$ of
Nature:
\[\limsup_{n \to \infty} \bar{R}_n^{ik} \leq 0 \quad \forall i,k \in \mathcal{I}, \quad \mathbb{P}_{\sigma,\tau}\mathrm{-as}.\]
\end{definition}

We introduce the following notations to define $\varepsilon$-internally consistency. Denote by $N_n(i)$
the set of stages before the $n$-th where the forecaster chose action $i$ and $\bar{\jmath}_n(i) \in \Delta(\mathcal{J})$ the empirical
distribution of Nature's actions on this set. Formally,
\begin{equation}\label{defNn}N_n(i)=\left\{m \in \{1,\ldots, n\};\ i_m=i \right\} \quad \mathrm{and} \quad
\bar{\jmath}_n(i)=\frac{\sum_{m \in N_n(i)} j_m}{|N_n(i)|} \in
\Delta(\mathcal{J}).\end{equation} A strategy is $\varepsilon$-internally
consistent if for every $i,k \in \mathcal{I}$
\[\limsup_{n \to \infty}\frac{|N_n(i)|}{n}\bigg(\rho\big(k,\bar{\jmath}_n(i)\big)-\rho\big(i,\bar{\jmath}_n(i)\big)- \varepsilon\bigg) \leq 0,\quad \mathbb{P}_{\sigma,\tau}\mathrm{-as}.\]

If we define, for every $\varepsilon \geq 0$, the $\varepsilon$-best
response correspondence by :
\[ \BR_{\varepsilon}(y)=\left\{x \in \Delta(\mathcal{I});\ \rho(x,y) \geq \max_{z \in \Delta(\mathcal{I})} \rho(z,y)-\varepsilon\right\},\]
then a strategy of the decision maker is $\varepsilon$-internally
consistent if any  action $i$ is either an $\varepsilon$-best
response to the empirical distribution of Nature's actions on
$N_n(i)$   or the frequency of $i$ is very small. We  will simply
denote $\BR_{0}$ by $\BR$ and  call it the best response
correspondence.

\medskip
From now on, given two sequences $\left\{l_m \in \mathcal{L}, a_m \in
\mathbb{R}^d;\ m \in \mathbb{N}\right\}$ where $\mathcal{L}$ is a finite set,
we will define the subset of integers $N_n(l)$ and the average
$\bar{a}_n(l)$ as in equation (\ref{defNn}).

\begin{proposition}[Foster \& Vohra \cite{FosterVohraCalibratedLearningCorrelatedEquilibrium}]\label{theointcons}
For every $\varepsilon \geq 0$, there exist $\varepsilon$-internally
consistent strategies.
\end{proposition}

Although the notion of internal regret is  a refinement of the notion of external regret (in the sense that any internally consistent strategy is also externally consistent), Blum \& Mansour \cite{BlumMansourJMLR} proved that  any externally consistent algorithm can be efficiently transformed into an internally consistent one (actually they obtained an even stronger property called \textsl{swap consistency}).

Foster \& Vohra \cite{FosterVohraCalibratedLearningCorrelatedEquilibrium} and Hart \& Mas-Colell
\cite{HartMasColellCorrelatedEquilibrium} proved directly the existence of
0-internally consistent strategies using different algorithms (with optimal rates and based
respectively on the Expected Brier Score and Blackwell's
approachability theorem). In some sense, we merge these two last proofs
 in order to provide a new one --- given in the following section ---
that can be extended quite easily to the partial monitoring
framework.

\subsection{A na\"ive algorithm, based on calibration}
The algorithm (a similar idea was used by Foster \& Vohra
\cite{FosterVohraCalibratedLearningCorrelatedEquilibrium}) that
constructs an $\varepsilon$-internally consistent strategy is based
on this simple fact: if the forecaster can, stage by stage, foresee
the law of Nature's next action, say $y \in \Delta(\mathcal{J})$, then he just has to choose
any best response to $y$  at the following stage. The continuity of
$\rho$ implies that the forecasts need not be extremely precise
but only up to some $\delta>0$.

Let $\{y(l);\ l\in \mathcal{L}\}$ be a $\delta$-grid
 of $\Delta(\mathcal{J})$ (\textsl{i.e.}\ a finite set such that for every $y \in \Delta(\mathcal{J})$ there exists $l \in \mathcal{L}$
such that $\| y-y(l)\| \leq \delta$) and  $i(l)$ be a best response to  $y(l)$, for every $l \in \mathcal{L}$. Then if $\delta$ is small enough:
\[ \| y - y(l)\| \leq 2\delta \Rightarrow i(l) \in \BR_{2\varepsilon}(y)\]

It is possible to construct a \textsl{good  sequence of forecasts} by
computing a calibrated strategy (introduced by Dawid
\cite{DawidWellCalibrated} and recalled in the following subsection
\ref{sectioncalib3}).

\subsubsection{Calibration}\label{sectioncalib3}

Consider a two-person repeated game $\Gamma_c$ where, at stage $n$, Nature
chooses the state of the world $j_n$ in a finite set $\mathcal{J}$ and a decision maker (that will be referred in this setting as a predictor) predicts it by choosing
$y(l_n)$ in $\mathcal{Y}=\{y(l);\ l\in \mathcal{L}\}$, a finite $\delta$-grid of
$\Delta(\mathcal{J})$ -- its  cardinality is denoted by $L$. As usual, a behavioral strategy $\sigma$  of the
predictor (resp.\ $\tau$ of Nature) is a mapping from the set of
finite histories $H=\bigcup_{n \in \mathbb{N}}\left(\mathcal{L} \times
\mathcal{J}\right)^n$ to $\Delta(\mathcal{L})$ (resp.\ $\Delta(\mathcal{J})$). We also denote
by $\mathbb{P}_{\sigma,\tau}$ the probability generated by the
pair $(\sigma,\tau)$ over $(\mathcal{H},\mathcal{A})$ the set of infinite
histories embedded with the cylinder topology.

\begin{definition}[Dawid \cite{DawidWellCalibrated}]
A strategy $\sigma$ of the predictor is calibrated (with respect to
$\mathcal{Y}=\{y(l);\ l\in \mathcal{L}\}$) if for every strategy $\tau$ of Nature,
$\mathbb{P}_{\sigma,\tau}$-as:
\[\limsup_{n \to \infty}\frac{|N_n(l)|}{n}\bigg(\left\| \bar{\jmath}_n(l)-y(l)\right\|^2-\left\| \bar{\jmath}_n(l)-y(k)\right\|^2\bigg) \leq 0, \quad \forall k,l \in \mathcal{L},\]
where $\|\cdot\|$ is the Euclidian norm of
$\mathbb{R}^{J}$.
\end{definition}
In words, a strategy is calibrated if for every $l \in \mathcal{L}$, the empirical distribution of
states, on the set of stages where $y(l)$ was predicted, is closer
to $y(l)$ than to any other $y(k)$ (  or the \textsl{frequency of} $l$, $|N_n(l)|/n$, is small).

Given a finite grid of $\Delta(\mathcal{J})$, the existence of calibrated
strategies has been proved by Foster \& Vohra
\cite{FosterVohraAsymptoticCalibration}   using either  the Expected Brier Score or a minmax theorem (actually this second argument  is acknowledged to Hart). We give here a construction, related but simpler than
the one of Foster and Vohra, due to Sorin \cite{SorinUnpublished}.

\begin{proposition}[Foster \& Vohra \cite{FosterVohraAsymptoticCalibration}]\label{theocalib}
For any finite grid $\mathcal{Y}$ of $\Delta(\mathcal{J})$, there exist
calibrated strategies with respect to $\mathcal{Y}$ such that for
every strategy $\tau$ of Nature:
\[
\mathbb{E}_{\sigma,\tau}\left[\max_{l,k \in
\mathcal{L}}\frac{|N_n(l)|}{n}\bigg(\left\|
\bar{\jmath}_n(l)-y(l)\right\|^2-\left\|
\bar{\jmath}_n(l)-y(k)\right\|^2\bigg)\right] \leq
O\left(\frac{1}{\sqrt{n}}\right).\]
\end{proposition}
\textbf{Proof.} Consider the  auxiliary game  where, at
stage $n \in \mathbb{N}$, the predictor (resp.\ Nature) chooses $l_n
\in \mathcal{L}$ (resp.\ $j_n \in \mathcal{J}$) and the vector payoff is the matrix $U_n
\in \mathbb{R}^{L \times L}$ where
    \[U_n^{lk}=\left\{\begin{array}{cc}
    \| j_n - y(l)\|^2 - \| j_n - y(k)\|^2 & \mathrm{if \ }  l=l_n\\
    0& \mathrm{otherwise.}
    \end{array}\right.\]
A strategy $\sigma$ is calibrated with respect to $\mathcal{L}$ if
$\bar{U}_n$ converges to the negative orthant.
Indeed for every $l,k \in \mathcal{L}$, the $(l,k)$-th coordinate of
$\bar{U}_n$ is
\begin{eqnarray*}\bar{U}_n^{lk}&=&\frac{|N_n(l)|}{n}\frac{\sum_{m  \in N_n(l)}\| j_m - y(l)\|^2 - \| j_m - y(k)\|^2}{|N_n(l)|}\\&=&\frac{|N_n(l)|}{n}\bigg(\| \bar{\jmath}_n(l) - y(l)\|^2 - \| \bar{\jmath}_n(l) - y(k)\|^2\bigg).\end{eqnarray*}

Denote by
$\bar{U}_n^+:=\left\{\max\left(0,\bar{U}_n^{lk}\right)\right\}_{l,k
\in \mathcal{L}}=:\bar{U}_n-\bar{U}_n^-$ the positive part of
$\bar{U}_n$ and by $\lambda_n \in \Delta(\mathcal{L})$ any invariant
measure of $\bar{U}_n^+$. We recall that $\lambda$ is an
invariant measure of a nonnegative matrix $U$ if, for every $l
\in \mathcal{L}$,
\[\sum_{k \in
\mathcal{L}}\lambda(k)U^{kl}=\lambda(l)\sum_{k \in \mathcal{L}}U^{lk}.\]
Its existence is a consequence of Perron-Frobenius Theorem, see
\textsl{e.g.}\ Seneta \cite{Seneta}.

\bigskip

Define  the strategy $\sigma$ of the predictor inductively as
follows. Choose arbitrarily $\sigma(\emptyset)$,  the law of the
first action and  at stage $n+1$, play accordingly to any invariant
measure of $\bar{U}_n^+$. We claim that this strategy is an
approachability strategy of the negative orthant of $\mathbb{R}^{L
\times L}$ because it satisfies Blackwell
\cite{BlackwellAnalogue}'s sufficient condition: \[\forall
n \in \mathbb{N}, \langle
\bar{U}_n-\bar{U}_n^-,\mathbb{E}_{\lambda_n}\left[U_{n+1}|j_{n+1}\right]-\bar{U}_n^-\rangle\leq0.\]
Indeed, for every possible $j_{n +1} \in \mathcal{J}$:
\begin{equation}\label{approachcond}  \langle \bar{U}_n^+,\mathbb{E}_{\lambda_n}\left[U_{n+1}|j_{n+1}\right]\rangle =0=\langle \bar{U}_n^+,\bar{U}_n^-\rangle,\end{equation}
where the second equality follows from the definition of positive
and negative parts.

Consider the first equality.  The $(l,k)$-th
coordinate of
$\mathbb{E}_{\lambda_n}[U_{n+1}|j_{n+1}]$ is $\lambda_n(l)\left(\left\| j_{n+1}-y(l)\right\|^2-\left\| j_{n+1}-y(k)\right\|^2\right)$, therefore the coefficient of $\| j_{n+1}-y(l) \|^2$ in the first term  is $\lambda_n(l)\sum_{k \in \mathcal{L}}\left(\bar{U}_n^+\right)^{lk}-\sum_{k \in \mathcal{L}} \lambda_n(k) \left(\bar{U}_n^+\right)^{kl}$. This equals 0 since $\lambda_n$ is an invariant measure of $\bar{U}_n^+$.

\bigskip

Blackwell \cite{BlackwellAnalogue}'s result also implies that
$\mathbb{E}_{\sigma,\tau}\left[\| \bar{U}_n^+ \| \right]
\leq 2M_nn^{-1/2}$ for any strategy $\tau$ of Nature where $M_n^2=
\sup_{m \leq n}\mathbb{E}_{\sigma,\tau}\left[\left\|
U_m\right\|^2\right]=4L$.$\hfill \Box$

\bigskip

Interestingly, the strategy $\sigma$ we constructed in this proof is actually internally consistent in the game with action spaces $\mathcal{L}$ and $\mathcal{J}$ and payoffs defined by $\rho(l,j)=-\|j-y(l)\|^2$.

    \begin{corollary}\label{corolcalib}
    For any finite grid $\mathcal{Y}$ of $\Delta(\mathcal{J})$, there exists
    $\sigma$, a calibrated strategy with respect to $\mathcal{Y}$, such that for every strategy $\tau$ of Nature, with
    $\mathbb{P}_{\sigma,\tau}$ probability at least $1-\delta$:
    \[ \max_{l,k \in \mathcal{L}}\frac{|N_n(l)|}{n}\bigg(\left\| \bar{\jmath}_n(l)-y(l)\right\|^2-\left\| \bar{\jmath}_n(l)-y(k)\right\|^2\bigg) \leq \frac{2M_n}{\sqrt{n}}+\Theta_n,\]
    \begin{eqnarray}\nonumber \mathrm{where} &\Theta_n=&\min\bigg\{\frac{v_n}{\sqrt{n}}\sqrt{2\ln\left(\frac{L^2}{\delta}\right)}+\frac{2}{3}\frac{K_n}{n}\ln\left(\frac{L^2}{\delta}\right),\frac{K_n}{\sqrt {n}}\sqrt{2\ln\left(\frac{L^2}{\delta}\right)}\bigg\};\\ \nonumber
    &M_n =& \sup_{m \leq n}\sqrt{\mathbb{E}_{\sigma,\tau}\left[\left\|
    U_m\right\|^2\right]}\leq3\sqrt{L};\\ \nonumber
    &v_n^2 =&\sup_{m \leq n}\sup_{l,k \in \mathcal{L}} \mathbb{E}_{\sigma,\tau}\left[\left|U_n^{lk}-\mathbb{E}_{\sigma,\tau}\left[U_n^{lk}\right]\right|^2\right]\leq 3;\\ \nonumber
    &K_n=&\sup_{m \leq n}\sup_{l,k \in \mathcal{L}}\left|U_n^{lk}-\mathbb{E}_{\sigma,\tau}\left[U_n^{lk}\right]\right|\leq 3.\end{eqnarray}
    \end{corollary}
\textbf{Proof.} Proposition \ref{theocalib} implies that
$\mathbb{E}_{\sigma,\tau}\left[\bar{U}_n\right]\leq
2M_nn^{-1/2}$. Hoeffding-Azuma's inequality  (see Lemma
\ref{HoeffdingAzumaInequalityAlgorithm} below in section \ref{subsectionHoeffdingAzumaa}) implies that with
probability at least $1 -\delta$ :
\[\bar{U}^{lk}_n - \mathbb{E}_{\sigma,\tau}\left[\bar{U}^{lk}_n\right]\leq \frac{K_n}{\sqrt {n}}\sqrt{2\ln\left(\frac{1}{\delta}\right)}.\]
Freedman's inequality (an analogue of Bernstein's inequality for
martingale see Freedman \cite{Freedman}, Proposition 2.1 or Cesa-Bianchi \& Lugosi \cite{CesaBianchiLugosi}, Lemma A.8) implies that with probability
at least $1-\delta$ :
\[\bar{U}^{lk}_n - \mathbb{E}_{\sigma,\tau}\left[\bar{U}^{lk}_n\right]\leq \frac{v_n}{\sqrt{n}}\sqrt{2\ln\left(\frac{1}{\delta}\right)}+\frac{2}{3}\frac{K_n}{n}\ln\left(\frac{1}{\delta}\right).\]
The result is a consequence of these two inequalities and of Proposition \ref{theocalib}. $\hfill \Box$

\medskip

The definition of $\Theta_n$ as a minimum (and the use of Freedman's
inequality) will be useful when we will refer to  this corollary
 in the subsequent sections. Obviously, in the current framework, $\Theta_n \leq \frac{3}{\sqrt{n}}\sqrt{2\ln\left(\frac{L^2}{\delta}\right)}$.

\subsubsection{Back to the Na\"ive Algorithm}\label{backnaive}
Let us now go back to the construction of $\varepsilon$-consistent
strategies in $\Gamma$. Compute $\sigma$, a
calibrated strategy with respect to a $\delta$-grid $\mathcal{Y}=\{y(l);\ l\in \mathcal{L}\}$ of $\Delta(\mathcal{J})$ in an abstract calibration
game $\Gamma_c$. Whenever the decision maker (seen as a
predictor) should choose the action $l$ in $\Gamma_c$, then he
(seen as a forecaster) chooses $i(l) \in \BR(y(l))$ in the original game~$\Gamma$. We claim that this defines a strategy
$\sigma_{\varepsilon}$  which is
$2\varepsilon$-internally consistent.

\begin{proposition}[Foster \& Vohra \cite{FosterVohraCalibratedLearningCorrelatedEquilibrium}]\label{naivefull}
For every $\varepsilon >0$, the strategy $\sigma_{\varepsilon}$
described above is $2\varepsilon$-internally consistent.
\end{proposition}
\textbf{Proof.} By definition of a calibrated strategy,
 for every $\eta >0$, there exists with probability~1, an integer $N
\in \mathbb{N}$ such that for every $l,k \in \mathcal{L}$ and for every $n
\geq N$ :
\[\frac{|N_n(l)|}{n}\bigg(\left\| \bar{\jmath}_n(l)-y(l)\right\|^2-\left\| \bar{\jmath}_n(l)-y(k)\right\|^2\bigg) \leq \eta.\]
Since $\{y(k);\ k \in \mathcal{L}\}$ is a $\delta$-grid of $\Delta(\mathcal{J})$, for
every $l \in \mathcal{L}$ and every $n \in \mathbb{N}$, there exists $k \in \mathcal{L}$
such that  $\left\|
\bar{\jmath}_n(l)-y(k)\right\|^2\leq \delta^2$, hence
$\left\| \bar{\jmath}_n(l)-y(l)\right\|^2 \leq
\delta^2+\eta\frac{n}{|N_n(l)|}$. Therefore, since $i(l) \in \BR(y(l))$:
\[ \frac{|N_n(l)|}{n} \geq \frac{\eta}{\delta^2} \Rightarrow \|\bar{\jmath}_n(l)-y(l)\|^2\leq 2 \delta^2 \Rightarrow \rho(k,\bar{\jmath}_n(l))-\rho(i(l),\bar{\jmath}_n(l))\leq 2 \varepsilon,\]
for every  $k \in \mathcal{I}$, $l \in \mathcal{L}$ and $n \geq N$. The $(i,k)$-th coordinate of
$\bar{R}_n$ satisfies:
\begin{eqnarray}\nonumber \frac{|N_n(i)|}{n}\bigg(\bar{R}_n^{ik}-2\varepsilon\bigg)&\leq&\frac{1}{n}\sum_{m  \in N_n(i)}\Big(\rho(k,j_m)-\rho(i,j_m)-2\varepsilon\Big)\\ \nonumber
&=&\frac{1}{n}\sum_{l : i(l)=i} \sum_{m \in
N_n(l)}\Big(\rho(k,j_m)-\rho(i,j_m)-2\varepsilon\Big)\\ \nonumber
&=&\sum_{l:
i(l)=i}\frac{|N_n(l)|}{n}\bigg(\rho(k,\bar{\jmath}_n(l))-\rho(i(l),\bar{\jmath}_n(l))-2\varepsilon\bigg).
\end{eqnarray}
Recall that either $\frac{|N_n(l)|}{n} \geq \frac{\eta}{\delta^2}$
and
$\rho(k,\bar{\jmath}_n(i))-\rho(i(l),\bar{\jmath}_n(l))-2\varepsilon
\leq 0$, or $\frac{|N_n(l)|}{n} < \frac{\eta}{\delta^2}$. Since
$\rho$ is bounded (by $M_{\rho}>0$), then :
\[\frac{|N_n(i)|}{n}\bigg(\bar{R}_n^{ik}-2\varepsilon\bigg) \leq \eta \frac{2M_{\rho}L}{\delta^2}, \quad \forall i \in \mathcal{I},\, \forall k \in \mathcal{I},\, \forall n \geq N,\]
which implies that $\sigma$ is $2\varepsilon$-internally
consistent.$\hfill \Box$

\begin{remark}\label{ratefull}
This na\"ive algorithm only achieves $\varepsilon$-consistency and Proposition
\ref{theocalib} implies that
\[\mathbb{E}_{\sigma,\tau}\left[\max_{i,k \in \mathcal{I}}\left(\bar{R}^{ik}_n-\varepsilon\right)\right]\leq O\left(\frac{1}{\sqrt{n}}\right).\]
The constants
depend drastically on $L$, which is in the current framework in the order of $\varepsilon^{J}$, therefore it is not possible to
obtain 0-internally consistency at the same rate with a classic
doubling trick argument (\textsl{i.e.}\ use a $2^{-k}$-internally
consistent strategy on $N_k$ stages, then switch to a
$2^{-(k+1)}$-internally consistent strategy, and so on, see e.g.\ Sorin \cite{SorinSupergames}, Proposition 3.2 page 56).

Moreover, since this algorithm is based on calibration, it computes at each stage
 an invariant measure of a non-negative matrix; this can be done,
using Gaussian elimination, with $O\left(L^3\right)$ operations, thus  this algorithm is far from being efficient (since its computational complexity is polynomial in $\varepsilon$ and exponential in $J$). There exist $0$-internally consistent algorithms, see \textsl{e.g.}\ the reduction of Blum \& Mansour \cite{BlumMansourJMLR}, that do not have this exponential dependency in the complexity or in the constants.

On the bright side, this algorithm  can be modified  to obtain 0-consistency at optimal rate; obviously, it will still not be efficient with full monitoring (see section \ref{sectionoptimalalgofull}). However, it has to be understood as a tool that can be easily adapted in order to exhibit, in the partial monitoring case, an optimal internal consistent algorithm   (see  section \ref{optimal}). And in that last framework, it is not clear that we can remove the dependency on $L$ (especially for the internal regret).
\end{remark}

\subsection{Calibration and  Laguerre diagram}
Given a finite subset of Vorono\"i sites $\{z(l) \in \mathbb{R}^d;\ l \in \mathcal{L}\}$, the $l$-th Vorono\"i cell $V(l)$,  or the
cell associated to $z(l)$, is the set of points closer to $z(l)$
than to any other $z(k)$:
\[V(l) = \left\{ Z \in \mathbb{R}^d;\ \left\| Z - z(l) \right\|^2 \leq \left\| Z - z(k) \right \|^2, \quad \forall k \in \mathcal{L}\right\},\]
where $\|\cdot\|$ is the Euclidian norm of $\mathbb{R}^d$. Each
$V(l)$ is a  polyhedron (as the intersection of a finite number of
half-spaces) and $\{V(l);\ l\in \mathcal{L}\}$ is a covering of $\mathbb{R}^d$.
A calibrated strategy with respect to $\{z(l);\ l\in \mathcal{L}\}$ has
the property that for every $l \in \mathcal{L}$,  the frequency of~$l$
goes to zero, or the empirical distribution of
states on $N_n(l)$, converges to $V(l)$.

The na\"ive algorithm uses the Vorono\"i diagram associated to an
arbitrary grid of $\Delta(\mathcal{J})$ and  assigns to every small cell  an $\varepsilon$-best reply to every point of
it; this is possible by continuity of $\rho$. A calibrated
strategy ensures that $\bar{\jmath}_n(l)$ converges to $V(l)$
(or the frequency of $l$ is small), thus choosing $i(l)$ on
$N_n(l)$ was indeed a $\varepsilon$-best response to
$\bar{\jmath}_n(l)$. With this approach, we cannot construct
immediately $0$-internally consistent strategy. Indeed, this would
require that for every $l \in \mathcal{L}$ there exists a 0-best response
$i(l)$ to every element $y$ in  $V(l)$. However,  there is no reason for them  to share a common best response because $\{z (l);\ l\in \mathcal{L}\}$ is chosen arbitrarily.

\bigskip

On the other hand, consider the simple game called \textsl{Matching
Penny}. Both players have two action $H$eads and $T$ails, so
$\Delta(\mathcal{J})=\Delta(\mathcal{I})=[0,1]$, seen as the probability of choosing
$T$. The payoff is 1 if both players choose the same action and -1
otherwise. Action $H$ (resp.\ $T$) is a  best response for Player 1
to any $y$ in $[0,1/2]$ (resp.\ in $[1/2,1]$). These two segments
are exactly the cells of the Vorono\"i diagram associated to
$\{y(1)=1/4,y(2)=3/4\}$, therefore, performing a calibrated strategy
with respect to $\{y(1),y(2)\}$ and playing $H$ (resp.\ $T$) on the
stages of type $1$ (resp.\ $2$) induces  a 0-internally consistent
strategy of Player 1.

\bigskip

This idea can be generalized to any game. Indeed, by Lemma \ref{decompbr} stated below, $\Delta(\mathcal{J})$ can be decomposed
into polytopial best-response areas (a polytope is the convex hull
of a finite number of points, its vertices). Given such a polytopial
decomposition, one can find a finer Vorono\"i diagram (\textsl{i.e.}\
any best-response area  is an union of Vorono\"i cells) and finally
use a calibrated strategy to ensure convergence with respect to this
diagram.

\medskip

 Although the construction of such a diagram is quite simple  in $\mathbb{R}$, difficulties arise in higher dimension -- even in $\mathbb{R}^2$. More importantly, the number of Vorono\"i sites can depend not only on the number of defining hyperplanes but also on the angles between them (thus being arbitrarily large even with a few hyperplanes). On the other hand, the description of a Laguerre diagram -- this concept generalizes Vorono\"i diagrams -- that refines a polytopial  decomposition is quite simple and is described in Proposition \ref{Voronoi} below. For this reason, we will consider from now on this kind of  diagram (sometimes also called Power diagram) .

Given a subset of Laguerre sites $\{z(l) \in
\mathbb{R}^d;\ l \in \mathcal{L}\}$ and weights $\{\omega(l) \in \mathbb{R};\ l
\in \mathcal{L}\}$,
 the $l$-th Laguerre cell $P(l)$ is defined by:
\[P(l) = \left\{ Z \in \mathbb{R}^d;\ \left\| Z - z(l) \right\|^2 - \omega(l) \leq \left\| Z -z(k) \right \|^2- \omega(k), \quad \forall k \in \mathcal{L}\right\},\]
where $\|\cdot\|$ is the Euclidian norm of $\mathbb{R}^d$.
Each $P(l)$ is a  polyhedron and $\mathcal{P}=\{P(l);\ l\in \mathcal{L}\}$ is a covering of
$\mathbb{R}^d$.

\begin{definition}
A covering $\mathcal{K}=\{K^i;\ i \in \mathcal{I}\}$ of a polytope $K$ with non-empty interior is a  polytopial  complex of $K$ if for every $i,j$ in the finite set $\mathcal{I}$, $K^i$
is a polytope with non-empty interior and the polytope $K^i \cap K^j$ has empty interior.
\end{definition} This definition extends naturally to a
polytope $K$ with empty interior, if we consider the affine subspace
generated by $K$.

\begin{lemma}\label{decompbr}
There exists a subset $\mathcal{I}' \subset \mathcal{I}$ such that $\{B^i;\ i \in \mathcal{I}'\}$ is a
 polytopial complex  of $\Delta(\mathcal{J})$, where $B^i$ is the $i$-th best response area
defined by
\[B^i = \{ y \in \Delta(\mathcal{J});\ i \in \BR(y)\}=\BR^{-1}(i).\]
\end{lemma}
\textbf{Proof.} For any $y \in \Delta(\mathcal{J})$, $\rho(\cdot,y)$ is linear
on $\Delta(\mathcal{I})$ thus it attains its maximum on  $\mathcal{I}$ and $\bigcup_{i \in
\mathcal{I}}B^i=\Delta(\mathcal{J})$. Without loss of generality, we can assume that
each $B^i$ is non-empty, otherwise we drop the index $i$. For every
$i,k \in \mathcal{I}$, $\rho(i,\cdot)-\rho(k,\cdot)$ is linear on $\Delta(\mathcal{J})$
therefore $B^i$ is a polytope; it is indeed  defined by
\begin{eqnarray}\nonumber B^i&=&\{y \in \Delta(\mathcal{J});\ \rho(i,y) \geq \rho(k,y),\, \forall k \in \mathcal{I}\}\\ \nonumber &=&\bigcap_{k \in \mathcal{I}} \{y \in \mathbb{R}^{J};\ \rho(i,y)-\rho(k,y)\geq 0\}\cap\Delta(\mathcal{J}),\end{eqnarray}
so it is the intersection of a finite number of half-spaces and the polytope $\Delta(\mathcal{J})$.

Moreover if $B_0^{ik}$, the interior of
$B^i \cap B^k$, is non-empty then $\rho(i,\cdot)$ equals
$\rho(k,\cdot)$  on the subspace generated by $B_0^{ik}$ and
therefore on $\Delta(\mathcal{J})$; consequently  $B^i=B^k$. Denote by $\mathcal{I}'$ any subset
of $\mathcal{I}$ such that for every $i \in \mathcal{I}$, there exists exactly one $i'
\in \mathcal{I}'$ such that $B^i=B^{i'}\neq\emptyset$, then  $\{B^i;\ i \in
\mathcal{I}'\}$ is a polytopial complex of $\Delta(\mathcal{J})$. $\hfill \Box$

\begin{proposition}\label{Voronoi}
Let $\mathcal{K}=\{K^i;\ i \in \mathcal{I}\}$ be a polytopial complex of a
polytope $K\subset \mathbb{R}^d$. Then there exists $\{z(l) \in
\mathbb{R}^d,\, \omega(l) \in \mathbb{R};\ l\in \mathcal{L}\}$, a finite set of
Laguerre sites and weights, such that the Laguerre diagram
$\mathcal{P}=\left\{P(l);\ l\in \mathcal{L}\right\}$ refines $\mathcal{K}$,
\textsl{i.e.}\ every $K^i$ is a finite union of cells.
\end{proposition}
\textbf{Proof.}  Let $\mathcal{K}=\{K^i;\ i\in \mathcal{I}\}$ be a polytopial
complex of $K \subset \mathbb{R}^d$. Each $K^i$ is a polytope, thus
defined by a finite number of hyperplanes. Denote by
$\mathcal{H}=\{H_t;\ t \in \mathcal{T}\}$ the set of all defining hyperplanes (the finite cardinality of $\mathcal{T}$ is denoted by $T$)
and $\widehat{\mathcal{K}}=\{\widehat{K}^l;\ l \in \mathcal{L}\}$ the finest
decomposition of $\mathbb{R}^d$ induced by $\mathcal{H}$ -- usually called
arrangement of hyperplanes -- which by definition refines $\mathcal{K}$. Theorem 3 and
Corollary 1 of Aurenhammer \cite{Aurenhammer} imply that
$\widehat{\mathcal{K}}$ is the Laguerre diagram associated to some
$\{z(l),\, \omega(l);\ l \in \mathcal{L}\}$ whose exact computation  requires the following notation:

\begin{itemize}
\item[i)]{for every $t\in \mathcal{T}$, let  $c_t \in \mathbb{R}^d$ and $b_t \in
\mathbb{R}$ (which can, without loss of generality, be assumed to be
non zero) such that
\[H_t= \left\{X \in \mathbb{R}^d;\ \langle X,c_t\rangle= b_t
\right\}.\]}
\item[ii)]{For every $l \in \mathcal{L}$ and $t \in \mathcal{T}$, $\sigma_t(l)=1$ if the origin of
$\mathbb{R}^d$ and $\widehat{K}^l$ are in the same halfspace defined
by $H_t$ and $\sigma_t(l)=-1$ otherwise.} \item[iii)]{For every $l
\in \mathcal{L}$, we define :
 \begin{equation}\label{defmuomega}z(l)= \frac{\sum_{t \in \mathcal{T}} \sigma_t(l)c_t}{T} \quad \mathrm{and} \quad  \omega(l)=\|z(l)\|^2+2\frac{\sum_{t \in \mathcal{T}}\sigma_t(l)b_t}{T}.\end{equation}}
\end{itemize}
Note that one can add the same constant to every  weight $\omega(l)$. $\hfill \Box$

\bigskip

Buck \cite{Buck} proved that  the number of cells defined by $T$
hyperplanes in $\mathbb{R}^d$ is bounded by $\sum_{k=0}^d{T \choose k}=:\phi(T,d)$, where ${T \choose k}$ is the binomial coefficient, \textsl{$T$ choose $k$}. Moreover, $T$ is
smaller than $I(I-1)/2$ (in the case where each $K^i$ has a
non-empty intersection with every other polytope), so $L \leq
\phi\left(\frac{I^2}{2}, d\right)$.

If $d \geq n$, then $\phi(n,d)=2^n$. Pascal's rule and a simple induction imply that, for every $n,d \in \mathbb{N}$, $\phi(n,d) \leq (n+1)^d$. Finally, for any $n \geq 2d$, by noticing that
\[ \frac{{n \choose d} +{n \choose d-1}+ \ldots+ {n \choose 0}}{{n \choose d}} \leq \sum_{m=0}^d \left(\frac{d}{n-d+1}\right)^m \leq \sum_{m=0}^{\infty} \left(\frac{d}{n-d+1}\right)^m\]
which equals $\frac{n-d+1}{n-2d+1}\leq 1+d$,
we deduce that $\phi(n,d) \leq (1+d){n \choose d} \leq (1+d)\frac{n^d}{d!}$.

\begin{lemma}\label{lemmadistcarre} Let $\mathcal{P}=\left\{P(l);\ l \in \mathcal{L} \right\}$ be a Laguerre diagram associated to the set of sites
and weights $\{z(l) \in \mathbb{R}^d,\, \omega(l) \in \mathbb{R};\ l
\in \mathcal{L}\}$. Then, there exists a positive constant $M_{P}
>0$ such that for every $Z \in \mathbb{R}^d$  if
\begin{equation}\label{equadistcarre2}\left\| Z - z(l) \right\|^2-\omega(l)  \leq \left\| Z - z(k)\right\|^2-\omega(k) + \varepsilon, \quad \forall l, k \in \mathcal{L}\end{equation}
then $d\left(Z,P(l)\right)$ is smaller than $M_{P}\varepsilon$.
\end{lemma}
The proof  can be found in Appendix \ref{sectionlemmadistcarre}; the constant $M_P$ depends on the Laguerre diagram, and more precisely
on the  inner products $\langle c_t, c_{t'} \rangle$, for
every $t,t'  \in \mathcal{T}$.
\subsection{Optimal algorithm with full monitoring}\label{sectionoptimalalgofull}
We reformulate Proposition
\ref{theocalib} and Corollary \ref{corolcalib}  in terms of
Laguerre diagram.
\begin{theorem}\label{theocalib2}
For any set of sites and weights $\{y(l) \in \mathbb{R}^{J},\,
\omega(l) \in \mathbb{R};\ l \in \mathcal{L}\}$ there exists a
 strategy $\sigma$ of the predictor such that for every strategy $\tau$ of Nature:
\[\mathbb{E}_{\sigma,\tau}\left[\left\|\left(\bar{U}_{\omega,n}\right)^+\right\|\right]\leq O\left(\frac{1}{\sqrt{n}}\right) \mathrm{\ where \ } U_{\omega,n} \mathrm{\ is \ defined \ by\ :}\]
\[U_{\omega,n}^{lk}=\left\{\begin{array}{ll}
\big[\| j_n - y(l)\|^2 -\omega(l)\big]- \big[\| j_n - y(k)\|^2-\omega(k) \big]& \mathrm{if \ }  l=l_n\\
0& \mathrm{otherwise}
\end{array}\right.\]
\end{theorem}

\begin{corollary}\label{corolcalib2}
For any set of sites and weights $\{y(l) \in \mathbb{R}^{J},\,
\omega(l) \in \mathbb{R};\ l \in \mathcal{L}\}$, there exists a
 strategy $\sigma$ of the predictor such that, for every strategy $\tau$ of Nature, with
$\mathbb{P}_{\sigma,\tau}$ probability at least $1-\delta$, and $l,l\in\mathcal{L}$:
\[\frac{|N_n(l)|}{n}\bigg(\left[\left\| \bar{\jmath}_n(l)-y(l)\right\|^2-\omega(l)\right]-\left[\left\| \bar{\jmath}_n(l)-y(k)\right\|^2-\omega(k)\right]\bigg) \leq \frac{2M_n}{\sqrt{n}}+\Theta_n\]
\begin{eqnarray}\nonumber \mathrm{where \ }  M_n&=& \sup_{m \leq n}\sqrt{\mathbb{E}_{\sigma,\tau}\left[\left\|
    U_{\omega,m}\right\|^2\right]}\leq4\sqrt{L}\|(b,c)\|_{\infty};\\
    \nonumber\Theta_n&=&\min\bigg\{\frac{v_n}{\sqrt{n}}\sqrt{2\ln\left(\frac{L^2}{\delta}\right)}+\frac{2}{3}\frac{K_n}{n}\ln\left(\frac{L^2}{\delta}\right),\frac{K_n}{\sqrt {n}}\sqrt{2\ln\left(\frac{L^2}{\delta}\right)}\bigg\};\\
    \nonumber v_n^2 &=&\sup_{m \leq n}\sup_{l,k \in \mathcal{L}} \mathbb{E}_{\sigma,\tau}\left[\left|U_{\omega,m}^{lk}-\mathbb{E}_{\sigma,\tau}\left[U_{\omega,m}^{lk}\right]\right|^2\right]\leq 4\| (b,c) \|_{\infty}^2;\\
    \nonumber K_n&=&\sup_{m \leq n}\sup_{l,k \in \mathcal{L}}\left|U_{\omega,m}^{lk}-\mathbb{E}_{\sigma,\tau}\left[U_{\omega,m}^{lk}\right]\right|\leq 4\|(b, c) \|_{\infty},\\
    \nonumber  \|(b,c)\|_{\infty}&=&\sup_{t \in \mathcal{T}}\| c_t\|+\sup_{t \in \mathcal{T}}|b_t|.\end{eqnarray}
Such a strategy is said to be calibrated with respect to $\{y(l),\, \omega(l);\ l \in \mathcal{L}\}$.
\end{corollary}
The proof are identical to the one of Proposition
\ref{theocalib} and Corollary \ref{corolcalib}. We have now the material to construct our new \textsl{tool algorithm}:
\begin{theorem}\label{regreinternnul}
There exists an internally consistent strategy $\sigma$ of the
forecaster such that for every strategy~$\tau$ of Nature and every
$n \in \mathbb{N}$, with $\mathbb{P}_{\sigma,\tau}$ probability
greater than $1-\delta$:
\begin{equation}
\max_{i,k \in \mathcal{I}} \bar{R}_{n}^{ik}\leq
O\left(\sqrt{\frac{\ln\left(\frac{1}{\delta}\right)}{n}}\right).
\end{equation}
\end{theorem}
\textbf{Proof.} The existence of  a Laguerre Diagram $\{Y(l);\ l \in \mathcal{L}\}$
associated to a finite set
 $\{y(l) \in \mathbb{R}^{J},\,
\omega(l) \in \mathbb{R};\ l\in \mathcal{L}\}$ that refines $\{B^i;\ i \in \mathcal{I}\}$ is implied by Lemma \ref{decompbr} and Proposition \ref{Voronoi}. So, for every $l \in \mathcal{L}$, there exists $i(l)$  such that  $Y(l) \subset B^{i(l)}$.  As in the na\"ive algorithm, the strategy $\sigma$ of the decision
maker is constructed through a  strategy $\widehat{\sigma}$
calibrated with respect to $\{y(l),\, \omega(l);\ l \in \mathcal{L}\}$. Whenever, accordingly to $\widehat{\sigma}$, the decision maker (seen as a
predictor) should play $l$ in
$\Gamma_c$, then he (seen as a forecaster) plays $i(l)$ in~$\Gamma$.

If we denote by $\widetilde{\jmath}_n(l)$ the projection of $\bar{\jmath}_n(l)$ onto $Y(l)$ then:
\begin{eqnarray*}
\bar{R}_{n}^{ik}&=&\sum_{l : i(l)=i} \frac{|N_n(l)|}{n}\bigg(\rho\big(k,\bar{\jmath}_n(l)\big)-\rho\big(i(l),\bar{\jmath}_n(l)\big)\bigg)\\
&\leq& \sum_{l : i(l)=i} \frac{|N_n(l)|}{n}\bigg(\bigg[\rho\big(k,\bar{\jmath}_n(l)\big)-\rho\big(k,\widetilde{\jmath}_n(l)\big)\bigg]\\
&& \qquad +\bigg[\rho\big(i(l),\widetilde{\jmath}_n(l)\big)-\rho\big(i(l),\bar{\jmath}_n(l)\big)\bigg]\bigg)\\
&\leq& \sum_{l : i(l)=i} \frac{|N_n(l)|}{n}\bigg(2M_{\rho}\left\|\widetilde{\jmath}_n(l)-\bar{\jmath}_n(l)\right\|\bigg)   \\
&\leq&  (2M_{\rho}M_PL)\max_{l, k \in
\mathcal{L}}\frac{|N_n(l)|}{n}\bigg(\left[\left\|\bar{\jmath}_n(l)-y(l)\right\|^2-
\omega(l)\right]\\
&& \quad -\left[\left\|\bar{\jmath}_n(l)-y(k)\right\|^2
-\omega(k)\right]\bigg)
\end{eqnarray*}
where the second inequality is due to the fact that  $i(l) \in
\BR(\widetilde{\jmath}_n(l))$ and the third to the fact that $\rho$
is $M_{\rho}$-Lipschitz. The fourth inequality is  a consequence of
Lemma \ref{lemmadistcarre}.

Corollary \ref{corolcalib2} yields that for every
strategy $\tau$ of Nature, with $\mathbb{P}_{\sigma,\tau}$
probability at least $1-\delta$:
\begin{eqnarray}\nonumber\max_{l,k}\frac{N_n(l)}{n}\bigg(\left[\left\|\bar{\jmath}_n(l)-y(l)\right\|^2-\omega(l)\right]-\left[\left\|\bar{\jmath}_n(l)-y(k)\right\|^2 -\omega(k)\right] \bigg)\leq \\ \nonumber \frac{8\sqrt{L}\| (b,c)\|_{\infty}}{\sqrt{n}} +\frac{4\|(b, c) \|_{\infty}}{\sqrt{n}}\sqrt{2\ln\left(\frac{L^2}{\delta}\right)},\end{eqnarray}
therefore with $\Omega_0=16M_{\rho}M_PL^{3/2}\| (b,c)\|_{\infty}$ and $\Omega_1=8M_{\rho}M_PL^{1/2}\|(b, c)\|_{\infty}$ one has that for every strategy of Nature and with probability at least $1-\delta$:
\[\max_{i,k \in \mathcal{I}} \bar{R}_{n}^{ik}=
\max_{i,k \in \mathcal{I}}
\frac{|N_n(i)|}{n}\bigg(\rho\big(k,\bar{\jmath}_n(i)\big)-\rho\big(i,\bar{\jmath}_n(i)\big)\bigg)
\leq \frac{\Omega_0}{\sqrt{n}}+\frac{\Omega_1}{\sqrt{n}}
\sqrt{2\ln\left(\frac{L^2}{\delta}\right)}.
\]
$\hfill \Box$
\begin{remark}
Theorem \ref{regreinternnul} is already well-known.  The
construction of this internally consistent strategy  relies on
Theorem \ref{theocalib2}, which is implied by the existence of
internally consistent strategies... Moreover, as mentioned before, it is far from being efficient since $L$ -- that enters both in the computational complexity and in the constant -- is polynomial in $I^{J}$. There exist efficient algorithms, see \textsl{e.g.}\ Foster \& Vohra \cite{FosterVohraCalibratedLearningCorrelatedEquilibrium} or Blum \& Mansour \cite{BlumMansourJMLR}.

 However,  the calibration  is defined in the space of Nature's action, where real payoffs are irrelevant; they are only used to decide which action is associated to each prediction. Therefore the algorithm
does not require that the forecaster observes his real payoffs, as long as he knows what is the best response to his information (Nature's action in this case). This
is precisely why our algorithm can be generalized to the partial
monitoring framework.\end{remark}

The polytopial decomposition of
$\Delta(\mathcal{J})$ induced by $\{b_t,\, c_t;\ t\in \mathcal{T}\}$ is exactly the same as
the one induced by $\{\gamma b(t),\, \gamma c(t);\ t \in \mathcal{T}\}$ for any
$\gamma>0$. Thus, by choosing $\gamma$ small enough, $\|
(b,c)\|_{\infty}$ --- and therefore the constants in
Corollary~\ref{corolcalib2} --- can be arbitrarily small
(\textsl{i.e.}\ multiplied by any $\gamma>0$).

However, these two Laguerre diagrams are associated  to the sets of
sites and weights $\mathcal{L}(1)$  and $\mathcal{L}(\gamma)$, where
$\mathcal{L}(\gamma)=\{\gamma z(l),\, \gamma \omega(l)+\gamma^2 \|
z(l)\|^2-\gamma\| z(l)\|;\ l \in \mathcal{L}\}$. If $\mathcal{L}(\gamma)$ is used instead of
$\mathcal{L}(1)$, then the constant $M_P$ defined in Lemma \ref{lemmadistcarre} should be
divided by $\gamma$. So, as expected, the constants in the proof of Theorem
\ref{regreinternnul} do not depend on $\gamma$.  From
now on, we will  assume that $\| (b,c)\|_{\infty}$ is smaller than 1.

\section{Partial monitoring}
\subsection{Definitions}
In the partial monitoring framework, the decision maker does not
observe Nature's actions. There is a finite set of signals $\mathcal{S}$ (of cardinality $S$) such that, at stage $n$ the forecaster receives only a random signal
$s_n \in \mathcal{S}$. Its law  is $s(i_n,j_n)$ where $s$ is a mapping from
$\mathcal{I} \times \mathcal{J}$  to $\Delta(\mathcal{S})$, known by the decision maker.

We define $\mathbf{s}$ from $\Delta(\mathcal{J})$ to $\Delta(\mathcal{S})^I$
by $\mathbf{s}(y)=\Big(\mathbb{E}_{y}\left[s(i,j)\right]\Big)_{i\in \mathcal{I}}
\in \Delta(\mathcal{S})^I$. Any element of $\Delta(\mathcal{S})^I$ is called a flag
(it is a vector of probability distributions over $\mathcal{S}$) and we will denote by $\mathcal{F}$ the range of $\mathbf{s}$.
Given a flag $f$ in  $\mathcal{F}$, the
decision maker cannot distinguish between any different mixed
actions $y$ and $y'$ in $\Delta(\mathcal{J})$ that generate $f$, \textsl{i.e.}\ such
that $\mathbf{s}(y)=\mathbf{s}(y')=f$. Thus $\mathbf{s}$ is the
maximal informative mapping about Nature's action. We denote by $f_n=\mathbf{s}(j_n)$ the (unobserved) flag of stage $n \in \mathds{N}$.

\begin{example}\label{examplelabel} Label efficient prediction (Example 6.8 in Cesa-Bianchi \& Lugosi \cite{CesaBianchiLugosi}):

Consider the following game. Nature chooses an outcome $G$ or $B$
and the forecaster can either observe the actual outcome (action
$o$) or choose to not observe it and pick a label $g$ or $b$.   His payoff is equal to  1 if  he
chooses the right label and otherwise is equal to 0. Payoffs
and laws of signals are defined by the
following matrices (where $a$, $b$ and $c$ are three different
probabilities over a finite given set $S$).
\begin{center}
\begin{tabular}{cc|c|c|cc|c|c|}
&\multicolumn{1}{c}{}& \multicolumn{1}{c}{$G$}&\multicolumn{1}{c}{$B$}&&\multicolumn{1}{c}{}&\multicolumn{1}{c}{$G$}&\multicolumn{1}{c}{$B$}\\
\cline{3-4}\cline{7-8}
&$o$&0&0&&$o$&$a$&$b$\\
\cline{3-4}\cline{7-8}
Payoffs: &$g$&0&1&\quad and signals: &$g$&$c$&$c$\\
\cline{3-4}\cline{7-8}
&$b$&1&0&&$b$&$c$&$c$\\
\cline{3-4}\cline{7-8}
\end{tabular}
\end{center}
Action $G$, whose best response is $g$, generates the flag $(a,c,c)$
and   action $B$, whose best response is $b$, generates the flag
$(b,c,c)$.  In order to distinguish between those two  actions, the
forecaster needs to know $s(o,y)$ although action $o$ is never a best response (but
is purely informative).
\end{example}

The worst   payoff compatible with $x$ and $f \in \mathcal{F}$ is defined by:
\begin{equation}
W(x,f)=\inf_{y \in \mathbf{s}^{-1}(f)}\rho(x,y),\end{equation} and $W$ is extended to $\Delta(\mathcal{S})^I$ by $W(x,f)=W\left(x,\Pi_{\mathcal{F}}(f)\right)$.

\medskip

As in the full monitoring case, we define, for every $\varepsilon
\geq 0$, the $\varepsilon$-best response multivalued mapping
$\BR_{\varepsilon} : \Delta(\mathcal{S})^I \rightrightarrows \Delta(\mathcal{I})$ by :
\[ \BR_{\varepsilon}(f)=\left\{x \in \Delta(\mathcal{I});\ W(x,f) \geq \sup_{z \in \Delta(\mathcal{I})}W(z,f)-\varepsilon\right\}.\]
Given a flag $f \in \Delta(\mathcal{S})^I$, the function $W(\cdot,f)$ may
not be linear so the best response of the forecaster might not
contain any element of $\mathcal{I}$.

\begin{example}Matching Penny in the dark:

Consider the Matching Penny game where the forecaster does not
observe the coin but always receives the same  signal $c$: every
choice of Nature generates the same flag $(c,c)$. For every $x \in [0,1]=
\Delta(\{H,T\})$ -- the probability of playing $T$ --, the worst compatible payoff $W(x,(c,c))=\min_{y
\in \Delta(J)}\rho(x,y)$ is equal to $-|1-2x|$ thus is non-negative
only for $x=1/2$. Therefore the only best response of the
forecaster is to play $\frac{1}{2}H+\frac{1}{2}T$, while actions $H$ and $T$ give
the worst payoff of -1.
\end{example}

The definition of external   consistency  and especially equation
(\ref{equadefeqregrext}) extend naturally to this framework:  a
strategy of the decision maker is externally consistent if he could
not have improved his payoff by knowing, before the beginning of the
game, the average flag:
\begin{definition}[Rustichini \cite{Rustichini}]
A strategy $\sigma$ of the forecaster is externally consistent  if
for every strategy $\tau$ of Nature:
\[ \limsup_{n \to +\infty} \max_{z \in \Delta(\mathcal{I})}W(z,\bar{f}_n)-\bar{\rho}_n \leq 0, \quad \mathbb{P}_{\sigma,\tau}\mbox{-as}.\]
\end{definition}

The main issue is the definition of internally consistency. In the
full monitoring case, the forecaster has  no internal
regret if, for every  $i \in \mathcal{I}$, the action  $i$ is a best-response
to the empirical distribution of Nature's actions, on the set of
stages where $i$ was actually chosen. In the  partial monitoring
framework, the decision maker's action should be a  best response to
the average flag. Since it might not belong to $\mathcal{I}$ but rather to $\Delta(\mathcal{I})$,  we will (following  Lehrer \& Solan \cite{LehrerSolanPSE})
distinguish the stages not as a function of the action actually
chosen, but as a function of its law.

We make an extra assumption on the characterization of the forecaster's strategy: it can be generated by a
finite family of mixed actions $\{x(l) \in \Delta(\mathcal{I});\ l\in \mathcal{L}\}$ such that, at stage
$n \in \mathbb{N}$, the forecaster chooses a type $l_n$ and, given
that type, the law of his action $i_n$ is $x(l_n) \in \Delta(\mathcal{I})$.

Denote by $N_n(l)=\{ m \in \{1,\ldots,n\};\ l_m = l\}$ the set of stages before the $n$-th whose type is $l$. Roughly speaking, a strategy will be $\varepsilon$-internally consistent
(with respect to the set $\mathcal{L}$) if, for every $l \in \mathcal{L}$, $x(l)$ is an $\varepsilon$-best response to $\bar{f}_n(l)$, the
average flag on $N_n(l)$ (or the frequency of the type $l$, $|N_n(l)|/n$, converges to zero).

The finiteness of $\mathcal{L}$ is required to get rid of  strategies that trivially insure that every frequency  converges to zero (for instance by choosing only once every mixed action).  The choice of $\{x(l);\ l\in \mathcal{L}\}$ and the description of the strategies are justified more precisely below by Remark \ref{remarkintrinsic} in
section \ref{optimal}.

\begin{definition}[ Lehrer \& Solan \cite{LehrerSolanPSE}]\label{defiregretpartial}
For every $n \in \mathbb{N}$ and every $l \in \mathcal{L}$, the average
internal regret of type $l$ at stage $n$ is
\[\mathcal{R}_n(l)=\sup_{x\in\Delta(\mathcal{I})}\left[W(x,\bar{f}_n(l))-\bar{\rho}_n(l)\right].\]

A strategy $\sigma$ of the forecaster is
$(\mathcal{L},\varepsilon)$-internally consistent if for every strategy $\tau$
of Nature:
\[ \limsup_{n \to +\infty} \frac{|N_n(l)|}{n}\bigg(\mathcal{R}_n(l) - \varepsilon\bigg) \leq 0, \quad \forall l \in \mathcal{L}, \quad \mathbb{P}_{\sigma,\tau}\mbox{-as}.\]
\end{definition}
In words, a strategy is $(\mathcal{L},\varepsilon)$-internally consistent if,
for every $l \in \mathcal{L}$, the forecaster could not have had, for sure,
a better payoff (of at least $\varepsilon$) if he had known, before
the beginning of the game, the average flag on $N_n(l)$ (or
the frequency of $l$ is  small).

\subsection{A na\"ive algorithm}\label{sectionnaivealgo}
\begin{theorem}[ Lehrer \& Solan \cite{LehrerSolanPSE}]\label{naivealgo}
For every $\varepsilon>0$, there exist
$(\mathcal{L},\varepsilon)$-internally consistent strategies.
\end{theorem}
 Lehrer \& Solan \cite{LehrerSolanPSE} proved the existence and constructed  such
strategies and an alternative, yet close, algorithm has been provided by
Perchet \cite{PerchetCalibrationALT}. The
main ideas behind them are  similar to the full monitoring
case so we will quickly describe them. For simplicity, we assume in the following sketch of the
proof, that  the decision maker fully observes the sequence of flags
$f_n=\mathbf{s}(j_n) \in \Delta(\mathcal{S})^I$.

Recall that $W$ is continuous (see Lugosi, Mannor \& Stoltz \cite{LugosiMannorStoltz}, Proposition
A.1), so for every $\varepsilon >0$ there exist two finite families
$\mathcal{G}=\{f(l) \in \Delta(\mathcal{S})^I;\ l\in \mathcal{L}\}$, a $\delta$-grid of
$\Delta(\mathcal{S})^I$, and $X=\{x(l) \in \Delta(I);\ l \in \mathcal{L}\}$ such that if
$f$ is $\delta$-close to $f(l)$ and $x$ is $\delta$-close to
$x(l)$ then $x$ belongs to $\BR_{\varepsilon}\left(f\right)$.  A calibrated algorithm
ensures that:
\begin{itemize}
\item[i)]{$\bar{f}_n(l)$ is asymptotically
$\delta$-close to $f(l)$ - because it is closer to $f(l)$ than
to every other $f(k)$;}
\item[ii)]{$\bar{\imath}_n(l)$ converges to $x(l)$ as soon as $|N_n(l)|$
is big enough - because  on $N_n(l)$ the  choices of action of the
decision maker are independent and  identically distributed
accordingly to $x(l)$;}
\item[iii)]{$\bar{\rho}_n(l)$ converges to $\rho(x(l),\bar{\jmath}_n(l))$ which is greater than $W\Big(x(l),\bar{f}_n(l)\Big)$ --- because $\bar{\jmath}_n(l)$ generates the flag $\bar{f}_n(l)$.}
\end{itemize}
Therefore, $W\Big(x(l),\bar{f}_n(l)\Big)$ is close to $W\Big(x(l), f(l)\Big)$ which is greater  than $W\Big(z,f(l)\Big)$ for any $z \in \Delta(\mathcal{I})$. As a consequence  $\bar{\rho}_n(l)$ is asymptotically greater (up to some $\varepsilon>0$) than $\sup_{z} W\Big(z,\bar{f}_n(l)\Big)$, as long as $|N_n(l)|$ is big enough.

\bigskip

The difference between the two algorithm lies in the construction of a calibrated strategy. On one hand, the algorithm of  Lehrer \& Solan \cite{LehrerSolanPSE} reduces to Blackwell's approachability of some convex set $\mathcal{C} \subset \mathds{R}^{LSI}$; it therefore requires to  solve at each stage a linear program of size polynomial in $\varepsilon^{SI}$, after a projection on $\mathcal{C}$. On the other hand, the algorithm of Perchet \cite{PerchetCalibrationALT} is based on the construction  given in section \ref{sectioncalib3}; it solves at each stage a system of linear equation of size also polynomial in $\varepsilon^{SI}$.

The conclusions of the full monitoring case also apply here: these highly non-efficient algorithms cannot be used directly  to construct
$(\mathcal{L},0)$-internally consistent strategy with optimal rates since the constants  depend drastically on $\varepsilon$ . We will rather prove  that one
can define wisely  once for all $\{f(l),\, \omega(l);\ l \in \mathcal{L}\}$ and $\{x(l);\ l \in \mathcal{L}\}$
(see Proposition \ref{propBR} and Proposition \ref{Voronoi}) so that
$x(l) \in \Delta(\mathcal{I})$ is a 0-best response to any flag $f$ in
$P(l)$, the Laguerre cell associated to $f(l)$ and $\omega(l)$.

The strategy associated with these choices will be  $(\mathcal{L},0)$-internally
consistent, with an optimal  rate of convergence  and a computational complexity polynomial in $L$.

\subsection{Optimal algorithms}\label{optimal}
As in the full monitoring framework (cf Lemma \ref{decompbr}), we
define for every $x \in \Delta(\mathcal{I})$  the $x$-best response area $B^x$
as the set of flags to which $x$ is a best response :
\[B^x = \left\{ f \in \Delta(\mathcal{S})^I;\ x \in \BR(f)\right\}=\BR^{-1}(x).\]
Since $W$ is continuous, the family $\left\{B^x;\ x \in
\Delta(\mathcal{I})\right\}$ is a covering of $\Delta(\mathcal{S})^I$. However,  one of
its finite subsets can be decomposed into  a finite  polytopial complex:
\begin{proposition}\label{propBR}
There exists a   finite family $X=\left\{x(l) \in \Delta(\mathcal{I});\ l \in
\mathcal{L}\right\}$ such that the family $\left\{ B^{x(l)};\ l \in \mathcal{L}\right\}$
of associated best response area can be further subdivided into a polytopial complex of
$\Delta(\mathcal{S})^I$.
\end{proposition}
The rather technical proof can be found in Appendix \ref{sectionpropBR}. In this framework and because of the lack of  linearity of $W$, any $B^{x(l)}$ might not be convex nor connected. However, each one of them is a finite union of polytopes and the family of all those polytopes is a complex of $\Delta(\mathcal{S})^I$.

\medskip

\begin{remark}\label{remarkintrinsic}
As a consequence of Proposition \ref{propBR}, there exists a finite set $X \subset \Delta(\mathcal{I})$  that
contains a best response to any flag $f$. In particular, if the
decision maker could observe the flag $f_n$ before choosing his
action $x_n$ then, at every stage, $x_n$ would be in $X$. So in the
description of the strategies of the forecaster, the finite set
$\{x(l);\ l\in \mathcal{L}\}=X$ is in fact intrinsic \textsl{i.e.}\ determined
by the description of the payoff and signal functions.
\end{remark}

As a consequence of this remark, mentioning $\mathcal{L}$ is irrelevant; so we will, from now on, simply speak of \textsl{internally consistent strategies}.

\subsubsection{Outcome dependent signals}
In this section, we assume that the laws of the signal received by
the decision maker are independent of his action. Formally, for every
$i, i' \in \mathcal{I}$, the two mappings $s(i,\cdot)$
and $s(i',\cdot)$ are equal. Therefore, $\mathcal{F}$ (the set of
realizable flags) can be seen as a polytopial subset of $\Delta(\mathcal{S})$.
Proposition \ref{propBR} holds in this framework, hence there
exists a finite family $\{x(l);\ l \in \mathcal{L}\}$ such that for any flag
$f \in \mathcal{F}$, there is some $l \in \mathcal{L}$ such that $x(l)$ is a
best-reply to $f$. Moreover, for a fixed $l \in \mathcal{L}$, the set of such
flags is a polytope.

\begin{theorem}\label{theooutcome}
There exists an internally consistent strategy $\sigma$ such
that for every strategy~$\tau$ of Nature, with
$\mathbb{P}_{\sigma,\tau}$-probability at least $1- \delta$:
\begin{equation}
\sup_{l \in \mathcal{L}}\frac{|N_n(l)|}{n}\mathcal{R}_{n}(l) \leq
O\left(\sqrt{\frac{\ln\left(\frac{1}{\delta}\right)}{n}}\right).
\end{equation}
\end{theorem}
\textbf{Proof.}  Propositions \ref{Voronoi} and
\ref{propBR} imply the existence of two finite families $\{x(l);\ l
\in \mathcal{L}\}$ and $\{f(l),\, \omega(l);\ l \in \mathcal{L}\}$ such that $x(l)$ is a
best response to any $f$ in $P(l)$, the Laguerre cell associated
to $f(l)$ and $\omega(l)$. Assume, for the moment, that for any
two different $l$ and $k$ in $\mathcal{L}$, the probability measures $x(l)$ and
$x(k)$ are different.

\bigskip

The strategy $\sigma$ is defined as follows. Compute a strategy
$\widehat{\sigma}$ calibrated with respect to $\{f(l),\,
\omega(l);\ l\in \mathcal{L}\}$. When the decision maker (seen as a predictor)  should choose $l \in \mathcal{L}$
accordingly to $\widehat{\sigma}$, then he (seen as a forecaster) plays accordingly to $x(l)$ in
the original game. Corollary \ref{corolcalib2} (with the assumption
that $\| (b,c) \|_{\infty}$
is smaller than 1) implies that with $\mathbb{P}_{\sigma,\tau}$
probability at least $1- \delta_1$:
\begin{eqnarray}\nonumber \max_{l \in \mathcal{L}}\frac{|N_n(l)|}{n}\bigg(\left[\left\| \bar{s}_n(l)-f(l)\right\|^2-\omega(l)\right]-\left[\left\| \bar{s}_n(l)-f(k)\right\|^2-\omega(k)\right]\bigg) \leq\\ \nonumber \frac{8\sqrt{L}}{\sqrt{n}}+\frac{4}{\sqrt{n}}\sqrt{2\ln\left(\frac{L^2}{\delta_1}\right)},\end{eqnarray}
therefore combined with Lemma \ref{lemmadistcarre}, this yields that :
\begin{equation}\label{equaout1} \max_{l \in \mathcal{L}} \frac{|N_n(l)|}{n}\left\|\bar{s}_n(l)-\widetilde{f}_n(l)\right\|\leq \frac{8M_P\sqrt{L}}{\sqrt{n}}+\frac{4M_P}{\sqrt{n}}\sqrt{2\ln\left(\frac{L^2}{\delta_1}\right)},\end{equation}
where $\widetilde{f}_n(l)$ is the projection of
$\bar{s}_n(l)$ onto $P(l)$.

\bigskip

Hoeffding-Azuma's inequality implies that with
$\mathbb{P}_{\sigma,\tau}$ probability at least $1-\delta_2$:
\begin{equation}\label{equaout2}
\max_{l \in
\mathcal{L}}\frac{|N_n(l)|}{n}\bigg\|\bar{s}_n(l)-\bar{f}_n(l)\bigg\|\leq
\sqrt{\frac{2\ln\left(\frac{2SL}{\delta_2}\right)}{n}}\end{equation}
and with probability at least $1-\delta_3$ :
\begin{equation}\label{equaout3}
\max_{l \in
\mathcal{L}}\frac{|N_n(l)|}{n}\bigg|\bar{\rho}_n(l)-\rho(x(l),\bar{\jmath}_n(l))\bigg|\leq
M_{\rho}\sqrt{\frac{2\ln\left(\frac{2L}{\delta_3}\right)}{n}}.
\end{equation}

$W$ is $M_{W}$-Lipschitz in $f$ (see Lugosi, Mannor \& Stoltz \cite{LugosiMannorStoltz}) and
$\mathbf{s}\left(\bar{\jmath}_n(l)\right)=\bar{f}_n(l)$ therefore:
\begin{equation}\label{equaout4}\bar{\rho}_n(l)\geq W\Big(x(l),\widetilde{f}_n(l)\Big)-\Big|\bar{\rho}_n(l)-\rho(x(l),\bar{\jmath}_n(l))\Big|-M_W\Big\|\bar{f}_n(l)-\widetilde{f}_n(l)\Big\|\\
\end{equation}
and $\max_{x \in \Delta(\mathcal{I})}W\left(x,\bar{f}_n(l)\right)$ is smaller than \begin{eqnarray}\label{equaout5}\nonumber \max_{x \in \Delta(\mathcal{I})}W\Big(x,\widetilde{f}_n(l)\Big)+M_W\left(\Big\|\bar{s}_n(l)-\bar{f}_n(l)\Big\|+\Big\|\bar{s}_n(l)-\widetilde{f}_n(l)\Big\|\right)\\
= W\Big(x(l),\widetilde{f}_n(l)\Big)+M_W\left(\Big\|\bar{s}_n(l)-\bar{f}_n(l)\Big\|+\Big\|\bar{s}_n(l)-\widetilde{f}_n(l)\Big\|\right)
\end{eqnarray}
since $x(l)$ is a best response to $\widetilde{f}_n(l)$. Equations
(\ref{equaout4}) and (\ref{equaout5}) yield
\begin{equation}\label{equaout6}
\mathcal{R}_{n}(l) \leq
2M_W\Big\|\bar{s}_n(l)-\bar{f}_n(l)\Big\|+2M_W\Big\|\bar{s}_n(l)-\widetilde{f}_n(l)\Big\|+\Big|\bar{\rho}_n(l)-\rho(x(l),\bar{\jmath}_n(l))\Big|.
\end{equation}
Combining equations (\ref{equaout1}), (\ref{equaout2}),
(\ref{equaout3}) and (\ref{equaout6}) gives that with probability at
least $1 - \delta$, if we define $\Omega_0=16M_PM_W\sqrt{L}$,
$\Omega_1=\left(2M_W+8M_WM_P+M_{\rho}\right)$ and
$\Omega_2=L\left(L+2S+2\right)$:
\begin{equation}
\sup_{l \in \mathcal{L}}\frac{|N_n(l)|}{n}\mathcal{R}_{n}(l) \leq
\frac{\Omega_0}{\sqrt{n}}+\frac{\Omega_1}{\sqrt{n}}
\sqrt{2\ln\left(\frac{2\Omega_2}{\delta}\right)}\end{equation}

\medskip

If there exist $l$ and $k$ such that $x(l)=x(k)$, then although the
decision maker made two different predictions $f(l)$ or $f(k)$,
he played accordingly to the same probability $x(l)=x(k)$. Define
$N_n(l,k)$ as the set of stages where the decision maker predicts either
$f(l)$ or $f(k)$ up to stage $n$, $\bar{f}_n(l,k)$ as the
average flag on this set, $\bar{\rho}_n(l,k)$ as the average
payoff and $\mathcal{R}_{n}(l,k)$ as the regret. Since $W(x,\cdot)$ is convex
for every $x \in \Delta(\mathcal{I})$, then $\max_{x \in \Delta(\mathcal{I})}W(x,\cdot)$
is also convex so $\frac{|N_n(l,k)|}{n}\max_{x \in \Delta(\mathcal{I})}W(x,\bar{f}_n(l,k))$ is smaller than
\[\nonumber \frac{|N_n(l)|}{n}\max_{x \in \Delta(\mathcal{I})}W(x,\bar{f}_n(l))+ \frac{|N_n(k)|}{n}\max_{x \in
\Delta(\mathcal{I})}W(x,\bar{f}_n(k))\]
\[\mathrm{and} \quad
-\frac{|N_n(l,k)|}{n}\bar{\rho}_n(l,k)=-\frac{|N_n(l)|}{n}\bar{\rho}_n(l)-\frac{|N_n(k)|}{n}\bar{\rho}_n(k)\]
so we still have \[\frac{|N_n(l,k)|}{n}\mathcal{R}_{n}(l,k) \leq
O\left(\sqrt{\frac{\ln\left(\frac{1}{\delta}\right)}{n}}
\right).\]
Hence the previous bound holds up to a factor $L$.$\hfill \Box$

\begin{remark}
Lugosi, Mannor \& Stoltz \cite{LugosiMannorStoltz} have constructed
an externally consistent strategy, \textsl{i.e.}\ such that,  asymptotically,
for any strategy $\tau$ of Nature:
\[ \bar{\rho}_n \geq \max_{z \in \Delta(\mathcal{I})} W\left(z,\bar{f}_n\right), \quad \mathbb{P}_{\sigma,\tau}\mathrm{-as}.\]
The final argument in the proof of Theorem \ref{theooutcome} also implies that  an internally consistent strategy is
also externally consistent, hence we can compare bounds between our
algorithm.

If the signals are deterministic, Lugosi, Mannor \& Stoltz  \cite{LugosiMannorStoltz}'s efficient algorithm has an   expected regret
smaller than $O\left(n^{-1/2}\right)$. However this bound became,
with random signals, $O\left(n^{-1/4}\right)$. Thus  our algorithm, along with  computing no internal regret, has a better
rate of convergence -- the optimal one. Concerning the computational complexity,  the true purpose of this algorithm being the minimization of internal regret, it is not efficient to bound external regret.
\end{remark}

\subsubsection{Action-Outcome dependant signals}
In this section, we consider the most general framework and we assume that the laws of the signals
might depend on the decision maker's actions. Our main result is the following:
\begin{theorem}\label{theoaction}
There exists an internally consistent strategy $\sigma$ such
that, for every strategy~$\tau$ of Nature, with
$\mathbb{P}_{\sigma,\tau}$ probability at least $1- \delta$:
\begin{eqnarray}
\max_{l \in \mathcal{L}}\frac{|N_n(l)|}{n}\mathcal{R}_{n}(l)\leq
O\left(\frac{1}{n^{1/3}}\sqrt{\ln\left(\frac{1}{\delta}\right)}+\frac{1}{n^{2/3}}\ln\left(\frac{1}{\delta}\right)\right).
\end{eqnarray}
\end{theorem}
\textbf{Proof.} The proof is essentially the same as the one of
Theorem \ref{theooutcome}, so we can assume  that $x(l) \neq x(k)$
for any two different $l$ and $k$ in $\mathcal{L}$. The only difference is due
to the fact that at stage $n \in \mathbb{N}$, the unobserved flag
$f_n$ has to be estimated (see \textsl{e.g.}\ Lugosi, Mannor \& Stoltz  \cite{LugosiMannorStoltz}).

\bigskip

Following Auer, Cesa-Bianchi, Freund \& Schapire \cite{AuerCesaBianchiFreundSchapire}, we define for every $l \in \mathcal{L}$ and $n \in \mathbb{N}$, the $\widehat{\gamma}_n$-perturbation  of $x(l)$ by $\widehat{x}(l,n) =
(1-\widehat{\gamma}_n)x(l) + \widehat{\gamma}_n u$ where $u$ is the
uniform probability over $\mathcal{I}$ and $(\widehat{\gamma}_n)_{n \in
\mathbb{N}}$ is a non-negative non-increasing sequence.  For every
$n \in \mathbb{N}$, let
\[e_n=\left(\frac{\mathds{1}_{i=i_n}}{\widehat{x}(l_n,n)[i_n]} \left(\mathds{1}_{s=s_n}\right)_{s \in \mathcal{S}}\right)_{i \in \mathcal{I}} \in \left(\mathds{R}^S\right)^I,\]
where $\widehat{x}(l_n,n)[i_n] \geq \gamma_n=\widehat{\gamma}_n/I>
0$  is  the  weight  put  by  $\widehat{x}(l_n,n)$
on  $i_n$. With this notation, $e_n$ is an unbiased
estimator of $f_n$ since
$\mathbb{E}_{\sigma,\tau}\left[e_n|h^{n-1}\right]=f_n$,  seen as an element of $\left(\mathds{R}^{S}\right)^I$.

\medskip

We define now the strategy of the forecaster. Assume that in an auxiliary game $\Gamma_c$, a predictor computes $\widetilde{\sigma}$, a calibrated
strategy  with respect to
$\{f(l),\, \omega(l);\ l \in \mathcal{L}\}$, but where the state at stage $n$
is the estimator $e_n \in \mathds{R}^{IS}$. When the decision maker (seen as a predictor) should choose $l_n$
 accordingly to $\widetilde{\sigma}$ in $\Gamma_c$, then he (seen as a forecaster) chooses $i_n$ accordingly to  $\widehat{x}(l_n)$ in the original game.

In order to use Corollary \ref{corolcalib2}, we need to bound $v_n$, $M_n$ and $K_n$.  In the current framework and thanks to Proposition \ref{Voronoi}, one has for every $l,k \in \mathcal{L}$ and $n \in \mathds{N}$:
 \[U_{\omega,n}^{l,k}=2\mathds{1}_{l=l_n} \sum_{t \in \mathcal{T}} \frac{\sigma_t(k)-\sigma_t(l)}{T} \bigg( \langle e_n, c_t \rangle +b_t\bigg),\]
 so using the fact that $\|(b,c)\|^2_{\infty} =1 $ and the definition of $e_n$:
\[ \sup_{l,k\in\mathcal{L}} \sup_{m \leq n} \mathds{E}_{\sigma,\tau}\left[\left|U_{\omega,m}^{l,k}\right|^2\right]\leq 16\mathds{E}_{\sigma,\tau}\Big[\|e_n\|^2\Big] \leq 16\sum_{i \in \mathcal{I}} \frac{\widehat{x}(l_n,n)[i]}{(\widehat{x}(l_n,n)[i])^2}\leq 16\frac{I}{\gamma_n}.\]
As a consequence, $K_n \leq 4\frac{1}{\gamma_n}$, $v_n \leq 4 \sqrt{\frac{I}{\gamma_n}}$ and $M_n \leq 4 \sqrt{\frac{LI}{\gamma_n}}$. Lemma \ref{lemmadistcarre} implies that, with $\mathds{P}_{\sigma,\tau}$ probability at least
$(1-\delta_1)$, for every $l \in \mathcal{L}$:
\[\frac{|N_n(l)|}{n}\left\|\bar{e}_n(l)-\widetilde{f}_n(l)\right\|\leq \frac{8\sqrt{LI}M_P}{\sqrt{\gamma_n n}}+\frac{8\sqrt{I}M_P}{\sqrt{\gamma_nn}}\sqrt{2\ln\left(\frac{L^2}{\delta_1}\right)}+\frac{8}{3}\frac{M_P}{\gamma_n n}\ln\left(\frac{L^2}{\delta_1}\right),\]
where $\widetilde{f}_n(l)$ is the projection of $\bar{e}_n(l)$ onto $P(l)$.

Following Lugosi, Mannor \& Stoltz  \cite{LugosiMannorStoltz}, since for every $i \in \mathcal{I}$ and $s \in \mathcal{S}$, $\mathds{E}_{\sigma,\tau}\left[|e^{i,s}_n|^2\right] \leq 1/\gamma_n$, Freedman's inequality   implies that with probability at least
$1-\delta_2$, for every $l \in \mathcal{L}$
\[
\frac{|N_n(l)|}{n}\Big\|\bar{e}_n(l)-\bar{f}_n(l)\Big\|
\leq
\sqrt{IS}\left(\sqrt{2\frac{1}{n\gamma_n}\ln\left(\frac{2LIS}{\delta_2}\right)}+\frac{2}{3n\gamma_n}\ln\left(\frac{2LIS}{\delta_2}\right)\right).\]
Hoeffding-Azuma's inequality  implies that  with probability at
least $1 - \delta_3$:
\[
\max_{l \in
\mathcal{L}}\frac{N_n(l)}{n}\Big|\bar{\rho}_n(l)-\rho(x(l),\bar{\jmath}_n(l))\Big|
\leq
M_{\rho}\sqrt{\frac{2}{n}\ln\left(\frac{2L}{\delta_3}\right)}+2M_{\rho}\frac{\sum_{m
\in N_n(l)}\widehat{\gamma}_m}{n},
\]
and by taking $\gamma_n=n^{-1/3}$, one has $\sum_{m \in N_n(l)}
\widehat{\gamma}_m \leq \frac{3I}{2}n^{2/3}$. As a consequence, for every $l \in \mathcal{L}$, with probability at least $1-\delta$:
\[
\frac{N_n(l)}{n}\mathcal{R}_{n}(l)\leq
\frac{\Omega_1}{n^{1/3}}+\frac{\Omega_2}{n^{1/3}}\sqrt{2\ln\left(\frac{2\Omega_5}{\delta}\right)}+\frac{\Omega_3}{n^{1/2}}\sqrt{2\ln\left(\frac{2\Omega_5}{\delta}\right)}+\frac{2}{3}\frac{\Omega_4}{n^{2/3}}\ln\left(\frac{2\Omega_5}{\delta}\right)
\]
with the constants defined by $\Omega_1=16M_PM_W\sqrt{LI}+3M_WM_{\rho}I$,
$\Omega_2=2M_W\sqrt{I}\left(8M_P+\sqrt{S}\right)$,  $\Omega_3=
M_{\rho}$, $\Omega_4=2M_W(4M_P+\sqrt{IS})$ and
$\Omega_5=L\left(L+2+2IS\right)$. They can be decreased if
concentration inequalities in Hilbert spaces are used (see section
\ref{Hilbert}).$\hfill \Box$

\medskip

In the label efficient prediction game defined in Example
\ref{examplelabel}, for every strategy $\sigma$ of the decision
maker there exists a sequence of outcomes such that the forecaster
expected regret is greater than $n^{-1/3}/7$ (see Theorem 5.1 in
Cesa-Bianchi, Lugosi \& Stoltz \cite{CesaBianchiLugosiStoltz}).
Therefore the rate of $n^{-1/3}$ of our algorithm is optimal
for both internal and external regret.

The computational complexity of this internally consistent algorithm is polynomial in $L$. Thus it can be seen, in some sense, as an efficient one.  A question left open is the existence of an algorithm whose computational complexity is polynomial in the minimal number of  best-response areas  required to cover $\Delta(\mathcal{S})^I$, see Proposition~\ref{propBR}.

 The following section \ref{optimal2} deals with a simpler question and exhibits an internally consistent algorithm which requires  to solve at each stage a linear program of size polynomial in $L_0$, the minimal number of polytopes on which $BR$ is constant, instead of a system of linear equations of size $L$.

\section{Concluding remarks}
\subsection{Second algorithm: calibration and polytopial complex.}\label{optimal2}

The algorithms we described are quite easy to run stage by stage
since the forecaster only needs to compute some invariant measures
of  non-negative matrices. However, they require to construct the
Laguerre diagram $\mathcal{P}=\left\{P(l);\ l \in \mathcal{L}\right\}$   given
the  set $\{b_t,\, c_t;\ t \in \mathcal{T}\}$. And we have shown that
$L$, which is a factor both in the
complexity of the algorithms and in their rate of convergence, can be in the order of $T^{SI}$ hence polynomial in $L_0^{SI}$.

\bigskip

This section is devoted to a modification of the algorithm that does
not require to compute a Laguerre diagram but which is more difficult, stage
by stage, to implement. The only difference between the two
algorithms is in the definition of calibration.

Let $\{K(l);\ l \in \mathcal{L}_0\}$ be a finite polytopial complex of $\Delta(\mathcal{J})$. It is defined by
two finite families $\left\{c_t \in \mathds{R}^{J},\, b_t \in
\mathds{R};\ t \in \mathcal{T}\right\}$ and $\left\{\mathcal{T}(l) \subset \mathcal{T};\ l \in \mathcal{L}\right\}$ such that:
\[K(l) = \left\{ y \in \Delta(\mathcal{J});\ \langle y, c_t \rangle \leq b_t,\, \forall t \in \mathcal{T}(l)\subset \mathcal{T}\right\}, \quad \forall l \in \mathcal{L}_0.\]
Let us define $(c_{t,l},b_{t,l})=(c_t,b_t)$ if $t \in \mathcal{T}(l)$ and $(c_{t,l},b_{t,l})=(0,0)$ otherwise. Then we can rewrite $K(l) = \left\{ y \in \Delta(\mathcal{J});\ \langle y, c_{t,l} \rangle \leq b_{t,l},\, \forall t \in \mathcal{T}\right\}$.

\begin{definition} A strategy $\sigma$  is calibrated w.r.t.\ the
complex $\{K(l);\ l\in \mathcal{L}_0\}$ if for every strategy $\tau$ of Nature,
$\mathds{P}_{\sigma,\tau}$-as:
\[\limsup_{n \to \infty}\frac{|N_n(l)|}{n}\bigg(\langle\bar{\jmath}_n(l),c_{t,l}\rangle-b_{t,l}\bigg) \leq 0, \quad \forall t \in \mathcal{T}, \forall l \in \mathcal{L}_0.\]
\end{definition}

\begin{theorem}\label{theocalibcompl}
There exist calibrated strategies w.r.t.\  any finite polytopial
complex $\{K(l);\ l \in \mathcal{L}_0\}$.
\end{theorem}
\textbf{Proof.} Consider the following  auxiliary two-person game
$\Gamma_c'$, where at stage $n \in \mathds{N}$ the predictor (resp.\
Nature) chooses $l_n \in \mathcal{L}_0$ (resp.\ $j_n \in \mathcal{J}$) which generates
the vector payoff $U_n \in \mathds{R}^{TL_0}$
defined by:
\[U_n^{lk}=\left\{\begin{array}{cc}
\langle \mathds{1}_{j_n=j},c_{t,l}\rangle-b_{t,l} & \mathrm{if\ } l=l_n\\
0& \mathrm{otherwise.}\end{array} \right.\]
Any strategy that  approaches the negative orthant $\Omega_-$ in $\Gamma_c'$ is calibrated w.r.t.\ the complex $\{K(l);\ l\in \mathcal{L}_0\}$.

\medskip

Blackwell's characterization of approachable convex sets (see
Blackwell \cite{BlackwellAnalogue}, Theorem 3) implies that  the
predictor can approach the convex set $\Omega_-$ if (and only if)
for every mixed action of Nature in $\Delta(\mathcal{J})$, he has an action $x
\in \Delta(\mathcal{L}_0)$ such that the expected payoff is in $\Omega_-$. Given $y_n \in \Delta(\mathcal{J})$, choosing  $l(y_n) \in
\mathcal{L}_0$, where $l(y_n)$ is the index of the polytope that contains
$y_n$, ensures that $\mathds{E}_{y_n,l(y_n)}[U_n]$ is in
$\Omega_-$. Therefore there exist calibrated strategies with respect
to any polytopial complex. $\hfill \Box$

\bigskip

This modification of the definition of calibration does not change
the other part of our algorithms nor the remaining of the proofs (in particular, to calibrate the sequence of unobserved flags, the forecaster must use $\widehat{\gamma}_n$-perturbations). The constants in the rates of convergence are now smaller since $L_0$ can be much smaller than $L$ and in $\Gamma_c'$, $\mathds{E}[\|U_n\|^2]$ is  bounded by
$O\left(\frac{T_0}{\gamma_n}\right)$ where $T_0=\sup_{l \in \mathcal{L}_0}T(l)$ is the maximum number of hyperplanes defining a polytope of the complex.

The main argument behind this algorithm (\textsl{i.e.}\ the characterization of approachable convex sets of Blackwell \cite{BlackwellAnalogue}) is quite close, in spirit, to the one of Lehrer \& Solan \cite{LehrerSolanPSE}. Note that however, with our representation, the projection on $\Omega_-$ can be computed linearly in $TL_0$, so polynomially in $L_0$. Therefore, it reduces to the construction of an approachability strategy and so -- as shown by Blackwell \cite{BlackwellAnalogue} -- to the resolution, at each stage, of a linear programming of size polynomial in $L_0$.

\subsection{Extension to the compact case}\label{sectioncompact}
We prove in this section that the finiteness of $\mathcal{J}$ is not required.

Assume that instead of choosing $j_n$ at stage $n \in \mathds{N}$ --
which generates the flag $f_n=\mathbf{s}(j_n)$ and an outcome
vector $\Big(\rho(i,j_n)\Big)_{i \in \mathcal{I}}$ -- Nature chooses
directly an outcome vector $O_n \in [-1,1]^I$ and a flag $f_n$
which belongs to $\mathbf{s}(O_n)$ where $\mathbf{s}$ is a
multivalued mapping from $[-1,1]^I$ into $\Delta(\mathcal{S})^I$. As before,
the decision maker's payoff is $O_n^{i_n}$ (the $i_n$-th coordinate
of $O_n$) and he receives a signal $s_n$ whose law is $f_n^{i_n}$.
Strategies of the forecaster and consistency are defined
as before.

\begin{theorem}\label{theocompact}
If the graph of  $\mathbf{s}$  is a polytope, then there
exists an internally consistent strategy $\sigma$ such that,
for every strategy $\tau$ of Nature, with $\mathds{P}_{\sigma,\tau}$
probability at least $1- \delta$:
\begin{eqnarray}
\max_{l \in \mathcal{L}}\frac{|N_n(l)|}{n}\mathcal{R}_{n}(l)\leq
O\left(\frac{1}{n^{1/3}}\sqrt{\ln\left(\frac{1}{\delta}\right)}+\frac{1}{n^{2/3}}\ln\left(\frac{1}{\delta}\right)\right).
\end{eqnarray}
\end{theorem}
The proof of this result is  identical to the one of Theorem \ref{theoaction}.

Note that the assumption that the graph of $\mathbf{s}$ is a polytope is fulfilled in the finite dimension case. The mapping $\mathbf{s}$ is multivalued since  in finite dimension there might exist two different mixed actions $y_1, y_1$ in $\Delta(\mathcal{J})$ that generate the same outcome vectore (\textsl{i.e.}\ $\rho(\cdot,y_1)=\rho(\cdot,y_2)=O$) but different flags (\textsl{i.e.}\ $f_1=\mathbf{s}(y_1)\neq \mathbf{s}(y_2)=f_2$). Hence we should have $f_1,f_2 \in  \mathbf{s}(O)$.
\subsection{Strengthening of the constants}\label{Hilbert}
We propose two different ideas to strengthen the constants of our
algorithm. First, we can use (as did Lugosi, Mannor \& Stoltz \cite{LugosiMannorStoltz}) only one concentration
inequality for every coordinate of the vector $U_{\omega,n}$ instead of one concentration inequality per coordinate. Second, we can implement sparser
vector payoffs (so that its norm decreases) by looking at a
slight different definition of calibration.

\subsubsection{Concentration Inequalities in Hilbert Spaces}\label{subsectionHoeffdingAzumaa}
The rates of convergence of our algorithms rely mainly on three
properties: Blackwell's approachability theorem, Hoeffding-Azuma's
and Freedman's inequalities. These tools allowed us to study the
convergence  of a sequence of vectors $\bar{U}_n^+$ towards 0.
Approachability is well defined for sequences of vectors, however
the two concentration inequalities  hold only for real valued
martingales. To circumvent this issue,  we used in the proofs the
fact that if a process $\left\{U_n \in \mathds{R}^{d}\right\}_{n \in
\mathds{N}}$ is a martingale then, for each coordinate, the process
$\left\{U_n^k \in \mathds{R}\right\}_{n \in \mathds{N}}$ is  a real
valued martingale. This does not use the fact that
$U_n$ might be  sparse and the use of concentration inequalities in
Hilbert space  can sharpen the constant.

\bigskip

Indeed, recall Hoeffding-Azuma's inequality:
 \begin{lemma}[Hoeffding\cite{Hoeffding}, Azuma \cite{Azuma}]\label{HoeffdingAzumaInequalityAlgorithm}
 Let $U_n$ be a sequence of martingale differences bounded by $K$, \textsl{i.e.}\ for every $n \in \mathds{N}$, $\mathds{E}_{\sigma,\tau}\left[U_{n+1}|h_{n}\right]=0$ and $|U_n|<K$.

Then for every $n \in \mathds{N}$ and  every $\varepsilon>0$:
\[
\mathds{P}_{\sigma,\tau}\left(\left|\bar{U}_n\right|\geq \varepsilon\right) \leq  2 \exp\left(\frac{-n\varepsilon^2}{2K^2}\right),
\]
which can be expressed as \begin{equation}\label{azumaeasy}\mathds{P}_{\sigma,\tau}\left(\left|\bar{U}_n\right|\leq K\sqrt{\frac{2}{n}\ln\left(\frac{2}{\delta}\right)}\right)\geq 1 - \delta.\end{equation}
\end{lemma}
Chen \& White \cite{ChenWhite} proved an equivalent property for vector martingale in $\mathds{R}^d$.

\begin{lemma}[Chen \& White \cite{ChenWhite}]
Let $U_n$ be a sequence of martingale differences in $\mathds{R}^d$ bounded
 almost-surely by $K>0$. Then  for every $n \in \mathds{N}$ and for every $\varepsilon>0$:
\[
\mathds{P}_{\sigma,\tau}\left(\left\|\bar{U}_n\right\|\geq \varepsilon\right) \leq  2\max\left\{1, \sqrt{\frac{n\varepsilon^2}{2K^2}}\right\} \exp\left(\frac{-n\varepsilon^2}{2K^2}\right)\leq 2 \exp\left(-\alpha\frac{n\varepsilon^2}{2K^2}\right),
\]
for every  $\alpha \leq 1- \frac{1}{2e}$ (which equals approximatively  $0.81$).
\end{lemma}

Assume that for every $n \in \mathds{N}$, $\|U_n\|_{\infty} \leq \|U\|_{\infty}$ and $\|U_n\|_2\leq \|U\|_2$; we can deduce from the use of  only  Hoeffding-Azuma's
inequality that:\[\mathds{P}_{\sigma,\tau}\left(\max_{l,k}\frac{|N_n(l)|}{n}\left|\bar{U}^{l,k}_n\right|\geq
\varepsilon\right) \leq  2L^2
\exp\left(\frac{-n\varepsilon^2}{2\|U\|_{\infty}^2}\right).\]However, Chen and
White's result, along with the fact that $\| U_n \|\leq
L$, implies that:
\[\mathds{P}_{\sigma,\tau}\left(\max_{l,k}\frac{|N_n(l)|}{n}\left|\bar{U}^{l,k}_n\right|\geq \varepsilon\right) \leq  2 \exp\left(\frac{-n\varepsilon^2}{4\|U\|_2^2}\right)\] which can reduce the dependency in $L$. The effects is even more dramatic when estimating the sequences of flags, since $e_n$ has only positive component (so $\|e_n\|_{\infty}=\|e_n\|_2$).

\bigskip

There also exist variants of Bernstein's inequality (see \textsl{e.g.}\ Yurinskii \cite{Yurinskii}) in Hilbert spaces that can be used in order to get more precise constants.

\subsubsection{Calibration with Respect of Neighborhoods}
\begin{definition}
Given a finite set $\mathcal{Y}=\{y(l) \in \mathds{R}^d,\, \omega(l) \in \mathds{R};\ l\in \mathcal{L}\}$,  $y(k)$ is a neighbor of $y(l)$ if $k \neq l$ and the dimension of  $P(l) \cap P(k)$ is equal to $d-1$.
\end{definition}

We defined a  calibrated strategy  with respect to $\mathcal{Y}$, as
a  strategy $\sigma$ such that $\bar{\jmath}_n(l)$ is asymptotically
closer to $y(l)$ than to any other $y(k)$ as soon as the
frequency of $l$ does not go  to zero. In fact, $\bar{\jmath}_n(l)$
needs only to be closer to $y(l)$ than to any of its neighbors. So
one can construct \textsl{neighbors}-calibrated strategies by
modifying the algorithm given in Proposition \ref{theocalib}; the
payoff at stage $n$ is now denoted by $U'_n$ and is defined by:
   \[\left(U'_n\right)^{lk}=\left\{\begin{array}{cc}
    \| j_n - y(l)\|^2 - \| j_n - y(k)\|^2 & \mathrm{if\ }  l=l_n \mathrm{\ and \ } k \mathrm{\ is \ a \ neighbor \ of \ } l\\
    0& \mathrm{otherwise}
    \end{array}\right.\]
The strategy consisting in choosing an invariant
measure of $\left(\bar{U}'_n\right)^+$ is calibrated and  $M^2_n=\sup_{m \leq n}
\mathds{E}_{\sigma,\tau}\left[\| U_m\|^2\right]$ equals
$4\mathcal{N}$, where $\mathcal{N}$ is the maximal number of
neighbors. This latter can be much smaller than $4$, and the gain from this
modification is limpid if  we consider $\varepsilon$-calibration.

Indeed, in order to  construct such
strategies, we usually take  any $\varepsilon$-discretization  of
$\Delta(J)$ so that $L=O\left(\varepsilon^{-(J-1)}\right)$.
However,  there exists a discretization such
that $\mathcal{N}=2^{-(J-1)}$, which is independent of
$\varepsilon$.

\appendix

\section{Proofs of technical results}\label{sectionalgoproof}
This section  is devoted to the proofs of previously mentioned results, \textsl{i.e.}\  Lemma \ref{lemmadistcarre} and Proposition \ref{propBR}.

\subsection{Proof of Lemma \ref{lemmadistcarre}}\label{sectionlemmadistcarre}
Let $l \in \mathcal{L}$ be fixed. we denote by $\mathcal{C}= \left\{ c_t \in \mathds{R}^d;\ t \in \mathcal{T}(l) \right\}$ the finite family of normal vectors to $(d-1)$-faces of $P(l)$ and by $\mathcal{B}=\left\{b_t \in \mathds{R};\ t \in \mathcal{T}(l)\right\}$ the family of scalars such that :
\[P(l)=\left\{ Z \in \mathds{R}^d;\ \langle Z, c_t\rangle \leq b_t,\, \forall t\in \mathcal{T}(l) \right\}.\]

Any points satisfying Equation (\ref{equadistcarre2}) belongs to
\[P_{\varepsilon}(l)=\left\{ Z \in \mathds{R}^d;\ \langle Z, c_t\rangle \leq b_t+\varepsilon,\, \forall t\in \mathcal{T}(l) \right\}.\]
For any vertex $v$ of $P(l)$, there exists $t_1,\ldots,t_d \in \mathcal{T}(l)$ such that
\[v=\bigcap_{k=1}^{d}\left\{Z \in \mathds{R}^d;\ \langle Z, c_{t_k} \rangle = b_{t_k}\right\}\]
and $\{c_{t_1},\ldots,c_{t_d}\}$ is a basis of $\mathds{R}^d$. If we denote by $v_{\varepsilon}$ the point defined by
\[v_{\varepsilon}=\bigcap_{k=1}^{d}\left\{Z \in \mathds{R}^d;\ \langle Z, c_{t_k} \rangle = b_{t_k} + \varepsilon\right\}\]
then $P_{\varepsilon}(l)$ is included in the convex hull of every $v_{\varepsilon}$.

Equation (\ref{equadistcarre2}) can be rephrased as: if $x$ belongs to  $P_{\varepsilon}(l)$ then $d(x,P(l))$ is smaller than $M_P\varepsilon$. Therefore it is enough to prove this property for every  $v_{\varepsilon}$ since  $d(\cdot,P(l))$ is a convex mapping  thus  maximized over a polytope on one of its  vertices.

\medskip

With these notations, for every $k \in \{1,\ldots,d\}$, $\langle
v_{\varepsilon}-v, c_{t_k}\rangle = \varepsilon$ and there exists
a unique decomposition $v_{\varepsilon}-v= \sum_{k= 1}^{d} \alpha_k
c_{t_k}$. Define the symmetric $d\times d$ Gram matrix $Q_l$ by
$Q^{kk'}_l=\langle c_{t_k},c_{t_{k'}}\rangle$ and
$\alpha=(\alpha_1,\ldots,\alpha_{d})$. Then  following classical
properties hold:
\begin{itemize}
\item[1)]{$\| v_{\varepsilon}-v\|^2 = \alpha^T Q_l \alpha$ and there exist  a $D=\diag(\lambda_1,\ldots,\lambda_{d})$ a diagonal matrix with $0<\lambda_1\leq \ldots \leq \lambda_{d}$ and a $d\times d$ matrix $P$ and  such that $P^{-1}=P^T$ and $Q_l=P^{T}DP$;}
\item[2)]{$Q\alpha=\underline{\varepsilon}=(\varepsilon,\ldots,\varepsilon)$ therefore $\alpha=Q_l^{-1}\underline{\varepsilon}$;}
\item[3)]{$\| v_{\varepsilon}-v \|^2=(Q_l^{-1}\underline{\varepsilon})^TQ_l(Q_l^{-1}\underline{\varepsilon})=\underline{\varepsilon}^TP^TD^{-1}P\underline{\varepsilon}\leq\varepsilon^2d\lambda_1^{-1}$.}
\end{itemize}
Therefore, for any $Z \in P_{\varepsilon}$ -- and in particular for any point that satisfies Equation (\ref{equadistcarre2}) --, $\| Z - \Pi_l(Z)\|\leq \max_{v} \|
v_{\varepsilon}-v\| \leq \varepsilon.\sqrt{d}\sqrt{\lambda_1}^{-1}$.  The result follows from the fact that $L$ is finite. The constant
$M_P$ in Lemma \ref{lemmadistcarre} is smaller than the square root of the inverse
of the smallest eigenvalue of all $Q_l$ times $\sqrt{d}$; it depends on the inner products $\langle c_t, c_{t'}\rangle$ and on the dimension of $\mathcal{F}$.

\subsection{Proof of proposition \ref{propBR}}\label{sectionpropBR}

\begin{definition}Let $K$  be a polytope. A correspondence $B: K \rightrightarrows \mathds{R}^d$ is  polytopial constant, if there exists $\{K(l);\ l\in \mathcal{L}\}$ a finite polytopial complex of $K$ and $\{x(l);\ l \in \mathcal{L}\}$ such that  $x(l) \in B(f)$ for every $f \in K(l)$.
\end{definition}
Let us now restate Proposition \ref{propBR}:
\begin{proposition}\label{propBRfinite}
$\BR$ is  polytopial constant.
\end{proposition}
This theorem is well-known and quite useful in the full monitoring case (see for example the Lemke-Howson \cite{LemkeHowson} algorithm). In the \textsl{compact case}, Proposition \ref{propBR} becomes:
\begin{proposition}\label{propBRcompact}
If $\mathbf{s}$ has a polytopial graph, then $\BR$ is  polytopial
constant.
\end{proposition}
The proofs of both propositions rely on polytopial parameterized
max-min programs defined in the next subsection.

\subsubsection{Constant Solution of a Polytopial Parameterized Max-Min Program}

A Polytopial Parameterized Max-Min Program (PPMP) is  defined as follows. Let $\mathcal{X}$ and  $\mathcal{Y}$  be two
Euclidian spaces of respective dimension $d_1$ and $d_2$. Consider the  program $(P_{f})$ - depending on a parameter $f$ that belongs to some polytope $\mathcal{F}$  in $\mathds{R}^{d_3}$ - that is defined by
\[ (P_{f}) : \quad \max_{\begin{array}{c}x \in \mathcal{X}\\s.t.\ Dx\leq d\end{array}} \min_{\begin{array}{c}y \in \mathcal{Y}\\s.t.\ E_{f}y\leq e_{f}\end{array}} xAy,\]
where $A$ is a $d_1\times d_2$ matrix,  $\{E_{f},\, e_{f};\ f \in \mathcal{F}\}$ is a family of  matrices and  vectors (we do not specify the sizes the matrices, as long as each inequality makes sense) and $D,d$ are also a fixed matrix and vector such that the admissible set $\mathcal{D}=\{x \in X;\ Dx\leq d\}$ is a polytope. The solution set of $(P_{f})$ is denoted by $B(f) \subset \mathcal{X}$ and this defines   a multivalued mapping $B(\cdot)$ from $\mathcal{F}$ into $\mathcal{X}$.

\begin{theorem}\label{lemmaBR2}
Assume that the  correspondence $S$ defined by:
\[
S : \begin{array}{ccl} \mathcal{F} &\rightrightarrows& \mathcal{Y}\\
f &\mapsto& S_{f}=\{y \in \mathcal{Y};\ E_{f}y \leq
e_{f}\}\end{array}
 \]
has a polytopial graph $\mathbf{S}$. Then  $B : \mathcal{F}\rightrightarrows \mathcal{X}$ is polytopial constant.
\end{theorem}
\textbf{Proof.}  Before going into full details, we first recall the following properties:
\begin{itemize}
\item[i)]{A linear program is minimized on a vertex of the polytopial feasible set (this is actually implied by the following point);}
\item[ii)]{Rockafella \cite{Rockafellar}, Theorem 27.4, page 270:
Given $x \in \mathcal{D}$ and $f \in \mathcal{F}$, if $y$ minimizes $xAy$ on   $S_{f}$ then
\[-xA \in \NC_{S_{f}}(y),\]
where $\NC_E(y)$ is the normal cone to the convex set $E \subset
\mathds{R}^d$ at $y\in E$ defined by :
\[\NC_E(y)=\left\{p \in \mathds{R}^d;\ \langle p, z - y \rangle,\, \forall z \in E\right\};\]}
\item[iii)]{Ziegler \cite{Ziegler}, Example 7.3, page 193: If $P$ is a polytope then  the finite family $\{\NC_{P}(v);\ v\, \mbox{ is a vertex of }\, P\}$ is a  polyhedral complex of $\mathds{R}^d$ called a normal fan (\textsl{i.e.}\ it is a finite family of polyhedra that cover $\mathds{R}^d$ and such that each pair has an intersection with empty interior);}
\item[iv)]{Billera \& Sturmfels \cite{BilleraSturmfels}, page 530: Since for every $f \in \mathcal{F}$, $S_f=\Pi^{-1}(f)$ where $\Pi: \mathbf{S} \subset \mathcal{F}\times \mathcal{Y} \to \mathcal{F}$ is the projection with respect to first coordinates, then there exists $\{K(l);\ l \in \mathcal{L}\}$, a polytopial complex of $\mathcal{F}$ such that the normal fan to $S_f$ is constant on every $K(l)$ (this can alternatively be deduced from the following point);}
\item[v)]{Rambau \& Ziegler \cite{RambauZiegler}, Proposition 2.4, page 221: On each of these polytopes $K(l)$, the mapping $f \mapsto S_f$ is linear. In particular, there exists a finite family of affine functions $Y(l)$ from $K(l)$ to $\mathcal{Y}$ such that the vertices of $S_f$ are exactly $\left\{y(f);\ y(\cdot) \in Y(l) \right\}$.}
\end{itemize}

Points i) and ii) imply that if $x_f$ maximizes $(P_f)$  -- which is then minimized at some a vertex of $S_f$ denoted by $y_f$,  because of point i) -- then it can be assumed that $-x_fA$ is a vertex of the polytope $NC_{S_f}(y_f) \cap \mathcal{D}_{A-}$ where $\mathcal{D}_{A-}:=\{-xA ;\ x \in \mathcal{D}\}$.  Thus $B(f)$, the  solution set to $(P_f)$ contains at least an element of
\[\mathbf{X}_f=\left\{x \in \mathcal{D}; -xA\, \mbox{ vertex of } \mathcal{D}_{A-} \cap NC_{S_f}(y_f), y_f\, \mbox{ vertex of } \, S_f \right\}.\]

By point iii), the normal fan and therefore $\mathbf{X}_f$ are constant on $K(l)$.  The latter can also be assumed to be finite by taking a unique representant $x \in \mathbf{X}_f$ for every vertices of the intersection of the normal fan and $\mathcal{D}_{A-}$. Since the number of different fans is finite, for any $f \in \mathcal{F}$, the solution set to $(P_f)$ contains at least an element of the finite set $\mathbf{X}=\bigcup_{f \in \mathcal{F}}\mathbf{X}_f$.

\medskip

Moreover, for every $\mathbf{x} \in \mathbf{X}$:
\begin{eqnarray*} B^{-1}(\mathbf{x})&=& \left\{ f \in \mathcal{F};\ \min_{y \in S_f} \mathbf{x}Ay \geq \max_{x' \in \mathcal{D}} \min_{y \in S_f}x'Ay\right\}\\
&=& \bigcup_{l \in \mathcal{L}} \left\{ f \in K(l);\ \min_{y \in S_f} \mathbf{x}Ay \geq \max_{x' \in \mathcal{D}} \min_{y \in S_f}x'Ay\right\}\\
&=& \bigcup_{l \in \mathcal{L}} \bigcap_{\mathbf{x'} \in \mathbf{X}}   \left\{ f \in K(l);\ \min_{y \in S_f} \mathbf{x}Ay \geq \min_{y \in S_f}\mathbf{x'}Ay\right\}\\
&=&\bigcup_{l \in \mathcal{L}} \bigcap_{\mathbf{x'} \in \mathbf{X}} \bigcup_{y'(\cdot) \in Y(l)} \left\{ f \in K(l);\ \min_{y \in S_f} \mathbf{x}Ay \geq \mathbf{x'}Ay'(f)\right\}\\
&=& \bigcup_{l \in \mathcal{L}}  \bigcap_{\mathbf{x'} \in \mathbf{X}} \bigcup_{y'(\cdot) \in Y(l)} \bigcap_{y(\cdot) \in Y(l)} \left\{ f \in K(l);\ \mathbf{x}Ay(f) \geq \mathbf{x'}Ay'(f)\right\},
\end{eqnarray*}
where, respectively, the second line is a consequence of point iv), the third line of the definition of $\mathbf{X}$ and the fourth and fifth lines of points i) and v).

\medskip

By point v), the two mapping $y(\cdot)$ and $y'(\cdot)$ are affine on $K(l)$, so each possible set
 \[\left\{ f \in K(l);\ \mathbf{x}Ay(f) \geq \mathbf{x'}Ay'(f)\right\}\]
 is a polytope as the intersection of an half-space and the polytope $K(l)$. Since, the intersection of a union of polytopes remains a union of polytopes, for every $\mathbf{x} \in \mathbf{X}$, $B^{-1}(\mathbf{x})$ is a finite  union of polytopes and $B$ is polytopial constant.  $\hfill \Box$

\bigskip

We can now prove simultaneously Propositions \ref{propBRfinite} and
\ref{propBRcompact}:

\subsubsection{Proof of Propositions \ref{propBRfinite} and
\ref{propBRcompact}}

Since $\mathbf{s}$ is linear, its graph,
denoted by $\mathbf{S}$, is a polytope. Theorem \ref{lemmaBR2} (with
$\mathcal{D} = \Delta(\mathcal{I})$) implies that  the solution, denoted by
$B(f)$ for every $f \in \mathcal{F}$, of the parameterized
program
\[ \max_{x \in \Delta(\mathcal{I})} \min_{y \in \mathbf{s}^{-1}(f)} \rho(x,y)\]
is polytopial constant. We denote by  $\{K(l);\ l \in \mathcal{L}\}$ a
corresponding  polytopial complex. If $B$ is
constant on $K(l)$, then it is also constant on
$\widehat{K}(l)=\Pi_{\mathbf{S}}^{-1}\left(K(l)\right)$, which is a
finite union of polytopes.$\hfill \Box$

\bigskip

\textbf{Acknowledgements:} I deeply thank my PhD advisor Sylvain
Sorin for its great help and support. I also acknowledge very useful
comments of Gilles Stoltz.

\end{document}